%% file: bare_jrnl_compsoc.tex
\documentclass[10pt,journal,twocolumn]{IEEEtran}
\ifCLASSOPTIONcompsoc
  \usepackage[nocompress]{cite}
\else
  \usepackage{cite}
\fi

\ifCLASSINFOpdf
\else
\fi

\usepackage{epsfig}
\usepackage{graphicx}
\usepackage{amsmath,amssymb}
\usepackage{eucal,bibspacing}
\usepackage{cite}
\usepackage{cs}
\usepackage{mathsymb}
\usepackage{colortbl}
\usepackage{color}
\usepackage{xcolor}

\usepackage{multirow}
\usepackage{pbox}
\usepackage{wrapfig}
\usepackage{caption}
\usepackage[linesnumbered,algoruled,boxed,lined]{algorithm2e}

\let\savedalgorithm\algorithm
\let\savedendalgorithm\endalgorithm

\usepackage{algorithm}

\def\bf{{\boldsymbol f}}

\begin{document}

\title{Pseudo-Pair based Self-Similarity Learning for Unsupervised Person Re-identification}

\author{Lin Wu, Deyin Liu*, Wenying Zhang, Dapeng Chen, Zongyuan Ge, Farid Boussaid, Mohammed Bennamoun, Jialie Shen
\IEEEcompsocitemizethanks{\IEEEcompsocthanksitem 
\protect\\ L. Wu is with The University of Western Australia, Perth 6009, Australia. E-mail: lin.wu@uwa.edu.au.
\protect\\ D. Liu (*corresponding author) is with Anhui Provincial Key Laboratory of Multimodal Cognitive Computation, School of Artificial Intelligence, Anhui University, Hefei 230039, China. E-mail: iedyzzu@outlook.com.
\protect\\ W. Zhang is with School of Information Engineering, Zhengzhou University, Zhengzhou 450001, China. E-mail: iewyzhang@zzu.edu.cn.
\protect\\ D. Chen is with Huawei, China. E-mail: chendapeng8@huawei.com.
\protect\\ Z. Ge is with Monash-Airdoc Research, Monash University, Melbourne, VIC 3000, Australia. E-mail: zongyuan.ge@monash.edu.
\protect \\ F. Boussaid and M. Bennamoun are with The University of Western Australia, Perth 6009, Australia. E-mail: \{Farid.Boussaid; Mohammed.Bennamoun\}@uwa.edu.au.
\protect\\ J. Shen is with Department of Computer Science, City, University of London, Northampton Square London EC1V 0HB, United Kingdom. E-mail: jerry.shen@city.ac.uk.
}
}

\IEEEtitleabstractindextext{%
\begin{abstract}
Person re-identification (re-ID) is of great importance to video surveillance systems by estimating the similarity between a pair of cross-camera person shorts. Current methods for estimating such similarity require a large number of labeled samples for supervised training. In this paper, we present a pseudo-pair based self-similarity learning approach for unsupervised person re-ID without human annotations. Unlike conventional unsupervised re-ID methods that use pseudo labels based on global clustering, we construct patch surrogate classes as initial supervision, and propose to assign pseudo labels to images through the pairwise gradient-guided similarity separation. This can cluster images in pseudo pairs, and the pseudos can be updated during training. Based on pseudo pairs, we propose to improve the generalization of similarity function via a novel self-similarity learning:it learns local discriminative features from individual images via intra-similarity, and discovers the patch correspondence across images via inter-similarity. The intra-similarity learning is based on channel attention to detect diverse local features from an image. The inter-similarity learning employs a deformable convolution with a non-local block to align patches for cross-image similarity. Experimental results on several re-ID benchmark datasets demonstrate the superiority of the proposed method over the state-of-the-arts.
\end{abstract}

\begin{IEEEkeywords}
Person re-identification, Pseudo pair construction, Unsupervised learning, Self-similarity learning.
\end{IEEEkeywords}}

\maketitle

\IEEEdisplaynontitleabstractindextext

\IEEEpeerreviewmaketitle

\section{Introduction}\label{sec:intro}

\IEEEPARstart{P}{erson} re-identification (re-ID) has received increasing attention in the computer vision community due to its significant role in video surveillance. Given a person shot as a query instance, the goal of re-ID is to find the corresponding subject from a large set of candidates by evaluating their visual similarities with respect to the query. Therefore, a reliable metric (or similarity) function is vital to an accurate person re-ID model. Supervised metric learning approaches have achieved remarkable success in jointly producing discriminative features and a precise metric function \cite{CAN,Part-Bilinear,Improved-re-ID,Dual-part-align}. However, they generally require considerable labelled data across camera pairs so as to learn an embedding that maps similar examples to nearby points while separating dissimilar examples far apart from each other. For this, these approaches exploit some objective functions in terms of pairwise \cite{CASCL,Improved-re-ID,Deep-re-ID}, triplet \cite{Defence-triplet-loss}, quadruplet \cite{Chen-quadruplet-cvpr2017}, and lifted structured loss \cite{Wu-CVIU-2018}. However, it requires expensive labelling to train the metric learning model with pairs \cite{Improved-re-ID,Deep-re-ID}, or triplets of images \cite{Defence-triplet-loss,Wu-TCSVT,WU-Video-Re-ID}. Moreover, by requiring a dataset with both identity and camera information, supervised approaches are inherently biased towards the dataset they are trained on and are therefore not generalizable. 

In contrast to supervised approaches, their unsupervised counterparts alleviate the need for annotation, which can be very costly to obtain in real-world scenarios. Several \textit{Unsupervised domain adaptation} (UDA) person re-ID methods \cite{SPGAN,PTGAN,Exemplar,MMT} aim to train a model on a labeled source dataset before adapting it to the unlabelled target dataset. However, preparing the labelled source dataset is costly while there is no measure to ascertain the domain gap between the source and the target datasets. \textit{Purely unsupervised methods} \cite{CASCL,BUC,Unsupervised-TIP20,MMCL,SSL,TJ-AIDL,PAUL,Radial-Distance-U-Re-ID} differ from UDA techniques in that they have no identity information whether from the source or target datasets. Such methods can perform unsupervised similarity learning by first generating pseudo labels for image clusters or individual images. They then improve the grouping of similar images within each identity. Though these approaches have been shown to yield state-of-the-art results, they raise two prominent concerns: \textbf{(i)} The clusters may be ill-generated with noisy labels, e.g., clustering image features extracted from a pre-trained network may result in imperfect clusters. The re-ID performance would be greatly degraded if images were assigned incorrect pseudo labels. Moreover, such methods need to estimate the global clustering of the whole training data because they apply soft-max formulation to assess the clustering membership for each image. This poses a strong constraint to a purely unsupervised re-ID setup. \textbf{(ii)} Discriminative features, such as body parts and textures, can improve the generalization of a re-ID model. However, these features have not been jointly learned with the similarity function.

In this work, we are interested in developing a pseudo-pair based self-similarity learning for purely unsupervised re-ID. For this, we need to generate pseudo pairs, and then perform fine-grained similarity learning to improve the generalization by learning discriminative features from the image itself. Unlike conventional methods that use pseudo labels based on global clustering, we explore the relative pseudo pairs based on the surrogate classes \cite{Exemplar-CNN}, i.e., the image gradients derived from a surrogate class can potentially describe the \textit{pairwise} relationship between images. This can naturally yield images in pseudo pairs without requiring the global clustering of training data. To derive the discriminative features which benefit the generalization ability of the similarity function, we investigate a few methods based on explainable attributes \cite{TJ-AIDL,MMCL,MMT}. These methods show that some attributes, such as outfit and textures, can be separated from identities and transferred to an unlabelled target dataset. However, our setup has no attribute annotations. Another recent work exploits the potential similarity (from the global body to local parts) within unlabeled images so as to build the identity-related groups \cite{SSG}. The core idea is based on patch matching, i.e., different parts can encode different discriminative features of an identity. Inspired by the aforementioned methods, we propose to discover discriminative features from the image itself and yet benefit the generalization of the similarity function.

\subsection{Our Approach}

We propose an unsupervised learning approach to learn a similarity function for person re-ID based on pseudo pairs. Instead of labelling images based on global clustering, we construct a set of surrogate classes as initial supervision. Each class consists of a seed patch and its variation patches formed by random transformations. Then we propose a gradient-guided similarity separation, which directly relates images to the corresponding identity manifold by measuring the discrepancy of image gradients derived w.r.t the surrogate class. This separation can be interpreted as a proxy for the pairwise comparison. It encourages the \textit{pairwise} clustering by measuring the discrepancy between gradient vectors. As a result, images are clustered in the form of pseudo pairs.

Given images in pseudo pairs, we further propose to improve the generalization of similarity function by learning more discriminative features from the image itself. In this sense, we propose to attend \textit{diverse} parts to make the parameters of similarity function robustly adaptive to \textit{unseen} shots. This mechanism is referred as intra-similarity learning, which is adversarially trained using channel attention without part annotations but with a regularization on parameter gradients to boost the generalization. This can effectively avoid the over-fitting on easy training samples. For example, if the upper body knowledge is sufficient to distinguish the training identities, the deep model will only focus on the body parts while ignoring other useful information. Apart from the intra-similarity learning for diverse features on each image, the similarity function should be convenient for pairwise image comparison. Thus, we propose a second mechanism, known as inter-similarity learning, which is carried out by aggressively searching for patch similarity across images. To align patch correspondence, we employ the deformable convolution \cite{Deform-Conv} which operates a dynamic offset estimator to learn patch offsets across images. More specifically, to estimate the dynamic offsets, we adopt a non-local block \cite{Non-local,SSEN} that performs patch-wise similarity matching. These two mechanisms lead to the proposed fine-grained self-similarity pipeline.

The contributions of this paper are summarized as follows: 1) We propose an effective unsupervised self-similarity learning approach for person re-ID based on pseudo pairs. The method aims to jointly learn both useful features and the similarity function; 2) A gradient-guided similarity separation is proposed to encourage pairwise clustering of gradient vectors, which can improve the pseudo labeling during training; 3) To improve the generalization of the learned metric, we propose a self-similarity learning which consists of a intra-similarity learning on each image to discover diverse features, and a inter-similarity learning across images to capture the cross-view similarity.

The rest of this paper is organized as follows. Section \ref{sec:related} reviews recent works related to our method. Section \ref{sec:method} details the proposed similarity learning approach, and in Section \ref{sec:exp} we conduct experiments to evaluate our method in comparison with state-of-the-art models. Finally, we conclude the paper in Section \ref{sec:con}.

\input{related}

\input{method}

\input{Experiment}

\input{Conclusion}

\section*{Acknowledgement}
This work was funded by Australian Research Council (Grants DP210101682 and DP210102674). This work was partially funded by Anhui Natural Science Foundation Anhui Energy Internet Joint Fund (No. 2008085UD07), Anhui Provincial Key Research and Development Project (No. 202104a07020029), and Natural Science Foundation of China (No. 61876002). Wenying Zhang was supported by Development of Standard Terminology Speech Recognition Technology based on Multi-Type Work of Comprehensive Practical Training on High-Speed Rail under No. 20200309A. 

\ifCLASSOPTIONcaptionsoff
  \newpage
\fi



\bibliographystyle{IEEEtran}\small
\bibliography{allbib}

%
%
%

%

\begin{IEEEbiography}[{\includegraphics[width=1in,height=1.25in,clip,keepaspectratio]{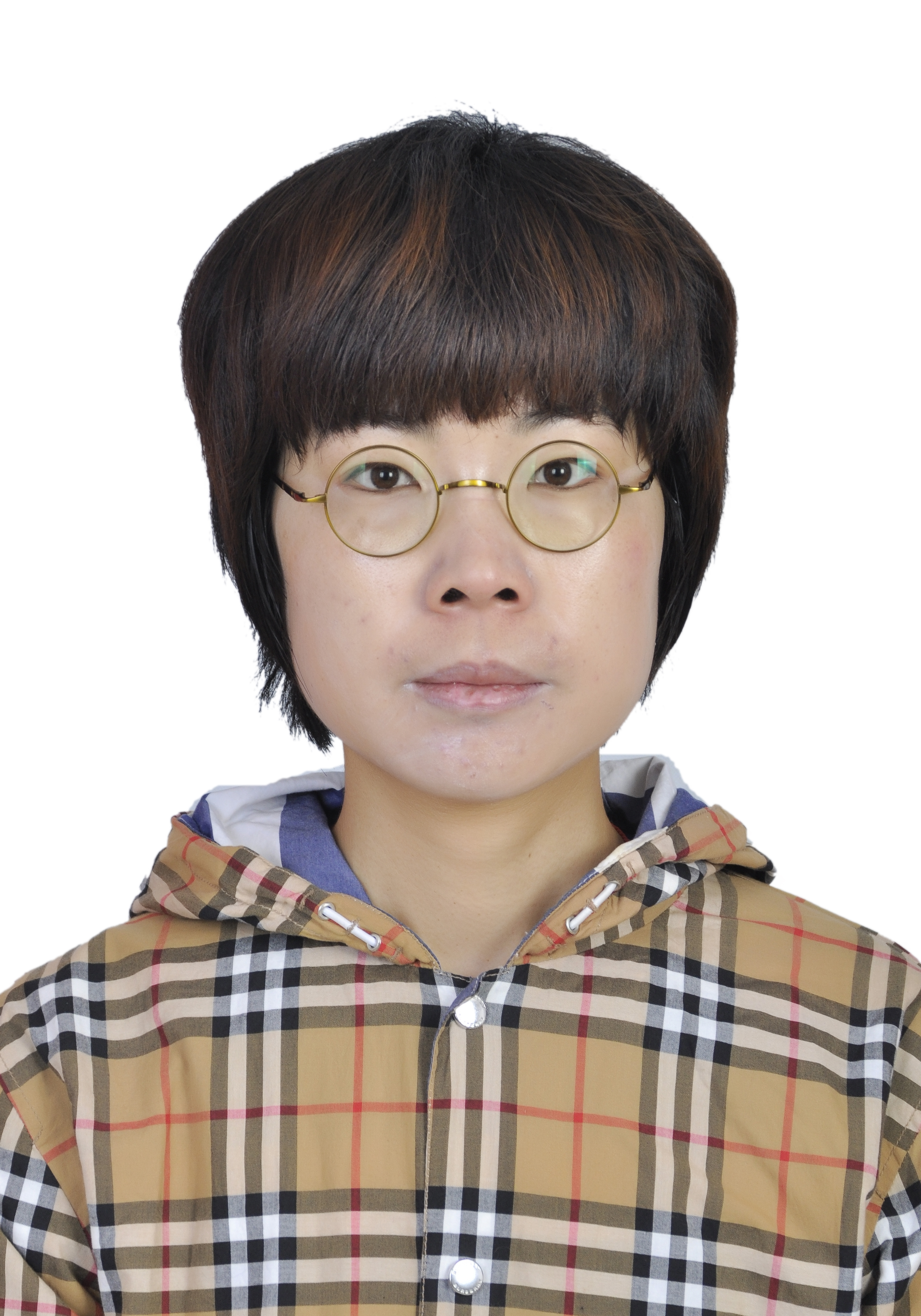}}]{Lin Wu} received a PhD from The University of New South Wales, Sydney, Australia in 2014. She is currently working as a senior research fellow with Department of Computer Science and Software Engineering, The University of Western Australia. She was previously working as a research fellow in University of Adelaide, University of Queensland, Australia and a tenure-track professor in Hefei University of Technology, China. She has intensively published 60+ peer-reviewed academic papers (including two book chapters) in premier journals and proceedings. She served as Area Chair with ACM Multimedia 2022. She is the recipient of the Award for Growth of 2021 The 4th Eureka International Innovation and Entrepreneurship Competition (Eureka IIEC 2021).
\end{IEEEbiography}

\begin{IEEEbiography}[{\includegraphics[width=1in,height=1.25in,clip,keepaspectratio]{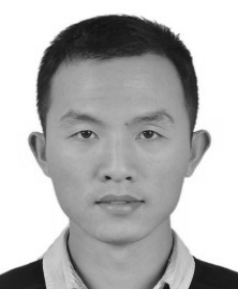}}] {Deyin Liu} received his B.E. degree from Zhengzhou University, China, in 2010. He is currently working with School of Artificial Intelligence, Anhui University, China. His main research interests include optimization in computer vision, unsupervised learning and sparse representation learning.
\end{IEEEbiography}

\begin{IEEEbiography}[{\includegraphics[width=1in,height=1.25in,clip,keepaspectratio]{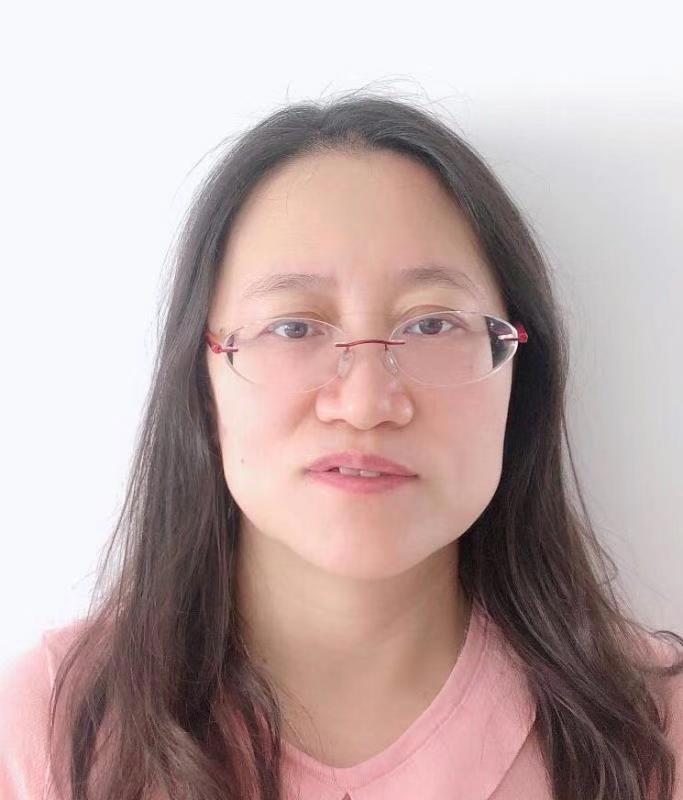}}] {Wenying Zhang} received the B.S. degrees in Electric traction and transmission control from Southwest Jiaotong University, Chengdu, China in 1994, and the M.S. degree in Information and communication systems from Zhengzhou University, Zhengzhou in 2002. Since 2002, she has been with School of Information Engineering at Zhengzhou University. Her research interests include pattern recognition, image and video processing and speech signal processing.
\end{IEEEbiography}

\begin{IEEEbiography}[{\includegraphics[width=1in,height=1.25in,clip,keepaspectratio]{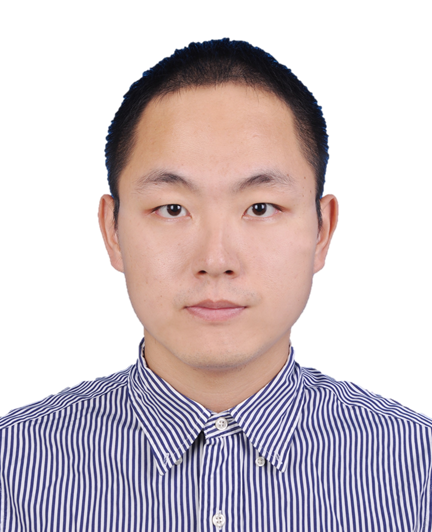}}]{Dapeng Chen} received his B.S. degree in biomedical engineering and a Ph.D. degree in control science and engineering from Xi'an Jiaotong University, China, in 2010 and 2016, respectively. He did postdoctoral research at the Multimedia Laboratory, The Chinese University of Hong Kong. He is currently a Research Scientist at Huawei. His research interests include computer vision, machine learning, and large-scale retrieval/clustering. 
\end{IEEEbiography}

\begin{IEEEbiography}[{\includegraphics[width=1in,height=1.25in,clip,keepaspectratio]{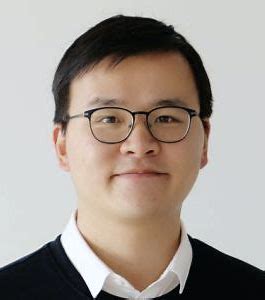}}]{Zongyuan Ge} is associate professor employed by Monash University with specific expertise in Medical AI development. He has a strong background in statistical analysis, machine learning and computer vision research. So far, he has published more than 30 peer-reviewed publications and patents. He has led or contributed to six international projects in the areas of dermatology, ophthalmology and radiology with major industry companies like IBM Watson Health, medical AI company Airdoc and medical service provider Molemap. 
\end{IEEEbiography}

\begin{IEEEbiography}[{\includegraphics[width=1in,height=1.25in,clip,keepaspectratio]{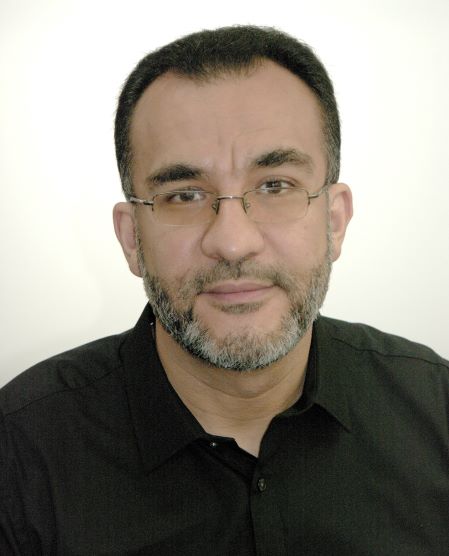}}]{Farid Boussaid} received the M.S. and Ph.D. degrees in microelectronics from the National Institute of Applied Science (INSA), Toulouse, France, in 1996 and 1999, respectively. He joined Edith Cowan University, Perth, Australia, as a Postdoctoral Research Fellow, and a Member of the Visual Information Processing Research Group, in 2000. He joined the University of Western Australia, Crawley, Australia, in 2005, where he is currently a Professor. His current research interests include neuromorphic engineering, smart sensors, and machine learning.
\end{IEEEbiography}

\begin{IEEEbiography}[{\includegraphics[width=1in,height=1.25in,clip,keepaspectratio]{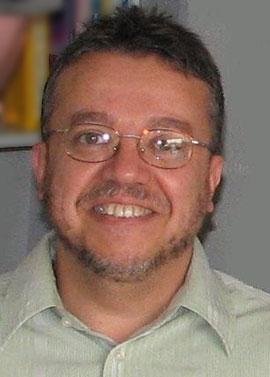}}]{Mohammed Bennamoun} (Senior Member, IEEE) is Winthrop Professor with the Department of Computer Science and Software Engineering, UWA and is currently a Researcher in computer vision, machine/deep learning, robotics, and signal/speech processing. He has published four books (available on Amazon, one edited book, one Encyclopedia article, 14 book chapters, 120+ journal articles, 250+ conference publications, 16 invited and keynote publications. His h-index is 54 and his number of citations is 12,500+ (Google Scholar). He was awarded 65+ competitive research grants, from the Australian Research Council, and numerous other Government, UWA and industry Research Grants. He successfully supervised 26+ the Ph.D. students to completion. He won the Best Supervisor of the Year Award at QUT, in 1998, and received award for research supervision at UWA, in 2008 and 2016, and Vice-Chancellor Award for mentorship, in 2016. He delivered conference tutorials at major conferences, including: IEEE Computer Vision and Pattern Recognition (CVPR 2016), Interspeech 2014, the IEEE International Conference on Acoustics Speech and Signal Processing (ICASSP) and European Conference on Computer Vision (ECCV). He was also invited to give a Tutorial at an International Summer School on Deep Learning (DeepLearn 2017).
\end{IEEEbiography}

\begin{IEEEbiography}[{\includegraphics[width=1in,height=1.25in,clip,keepaspectratio]{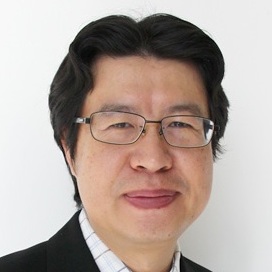}}]{Jialie Shen}  (Senior Member, IEEE) is a Professor in Computer Vision and Machine Learning, Department of Computer Vision, City, University of London. He received his PhD in Computer Science from the University of New South Wales (UNSW), Australia in the area of intelligent media search with large scale neural network. His research interests spread across subareas in artificial intelligence (AI), including computer vision, deep learning, image/video analytics, and machine learning. Professor Shen’s research results have expounded in 150+ publications at prestigious journals and conferences, with several awards: Lee Foundation Fellowship for Research Excellence Singapore, Microsoft Mobile plus Cloud Computing Theme Research Program Award, Best Paper Runner-up for IEEE Transactions on Multimedia (IEEE T-MM), Best Reviewer Award for Information Processing and Management (IP\&M) 2019 and ACM Multimedia 2020, and Test of Time Reviewer Award for Information Processing and Management (IP\&M) 2022. He also serves as the Associate Editor and (or) member of editorial board of leading journals: Information Processing and Management (IP\&M), Pattern Recognition (PR), IEEE Transactions on Circuits and Systems on Video Technology (IEEE T-CSVT), IEEE Transactions on Multimedia (IEEE T-MM), and ACM Transactions on Multimedia Computing, Communications, and Applications (ACM TOMM). He is a senior member of IEEE.
\end{IEEEbiography}



\end{document}

%% file: related.tex
\section{Related Work}\label{sec:related}
In this section, we review recent works with relevance to unsupervised learning, domain adaptation in person re-ID as well as self-similarity learning and deep attention models.

\subsection{Unsupervised Learning in Person Re-ID}
Conventional unsupervised methods usually can be categorized into three streams: designing hand-craft features \cite{LOMO,Gaussian}, exploiting localized salience statistics \cite{U-Re-ID-salience-match,U-Re-ID-salience}, or dictionary-learning based methods \cite{L1-graph}. However, these methods are under-performed due to the difficulty of designing features under camera changes and different illumination conditions. Recently, unsupervised domain adaptation (UDA) methods \cite{TJ-AIDL,PTGAN,Exemplar,SSL,SyRI,Self-paced-CL,MMT} were proposed to learn a re-ID model from a labelled source domain and an unlabelled target domain. For example, Wang et al. \cite{TJ-AIDL} proposed to learn an attribute-semantic and identity-discriminative representation from the source dataset, and transfer to the target domain. Zhong et al. \cite{Exemplar} proposed a framework that consisted of a classification module and an exemplar memory module for the labeled source data and the unlabelled target data. These methods take advantage of the external source domain, in which some cross-camera identities should be annotated. To alleviate the requirement on cross-camera identities, clustering-generated pseudo labels are employed in UDA methods \cite{Unsupervised-TIP20,MMT,Self-paced-CL,SDA-MMT}. However, the pseudo labels derived from clusters could be noisy. To overcome this issue, a Mutual-Mean-Teaching framework \cite{MMT} was presented to refine the quality of pseudo labels via on-line soft pseudo labels. Unlike these methods, we focus on fully unsupervised re-ID without any external source or identity annotations.

Purely unsupervised learning in person re-ID can learn a similarity through the relationship between individual images. \cite{CASCL,BUC,Unsupervised-TIP20,MMCL,SSL,TJ-AIDL,PAUL,Radial-Distance-U-Re-ID}. For example, \cite{Unsupervised-TIP20} introduced an unsupervised style transfer framework to generate style-transferred images for each identity, and then learned the similarity from both the original and transferred data. Another paradigm adopted an iterative clustering based deep learning \cite{SSL,BUC}, where the network was trained based on the clustering generated pseudo labels. As clustering based methods roughly divide images into clusters for training, they would make the model highly dependent on the clustering results, which could hinder the performance due to the inevitable noisy labels (e.g., images from different identities can be clustered into the same cluster). Instead of using the global clustering results as supervision guidance, we construct surrogate classes to cluster images into pseudo pairs.

\subsection{Domain Adaptation in Person Re-ID}
Domain adaptation can be applied into the re-ID task to align cameras under cross-view changes \cite{PTGAN,SPGAN,VCFL,Camera-style-re-ID}. For instance, Zheng et al. \cite{Camera-style-re-ID} addressed the cross-view situation by learning a camera-invariant descriptor subspace, which is known as camera-style adaptation. Deng et al. \cite{SPGAN} adopted domain adaptation to achieve image translation while maintaining the discriminative cues contained in the label space. These methods performed domain alignment in the feature space, without operating in the similarity space. The view confusion feature learning method (VCFL) \cite{VCFL} was recently introduced to learn view-agnostic identity-wise features via a combination of view-generic and view-specific models. However, VCFL \cite{VCFL} over-emphasizes the common parts (average images) while inattentively suppressing different views, since the adversarial training was likely to become optimal when attaining the averaged values so as to cater for various views. All these methods deal with cross-camera distribution by using domain adaptation. This setting is different from our problem.

\subsection{Self-Similarity Learning}
Self-similarity describes a relational structure of individual image features by computing the similarities between them \cite{Local-SS,SELFYNet-TSM,SSS}. For instance, Kwon et al. \cite{SELFYNet-TSM} proposed a robust motion representation with spatial-temporal self-similarities where each local region is represented as similarities to its neighbors in space and time. In person re-ID task, only a few methods have been so far developed based on self-similarity learning \cite{SSG,Radial-Distance-U-Re-ID}. For example, Seth et al. \cite{Radial-Distance-U-Re-ID} assigned the same pseudo label to each image and a set of its pose-transformed versions. This is done in conjunction with a metric learning loss to learn a latent space where samples belonging to different identities are located further apart than those belonging to the same identity. Another recent unsupervised self-similarity grouping (SSG) was presented to mine the potential similarities characterized by the training samples from a global to local manner \cite{SSG}. The core idea of SSG is based on patch matching, where different parts contain different discriminative information of a person. In contrast, our method differs from the above methods in two aspects: \textbf{1)} we do not require any data augmentation in the similarity learning process, and \textbf{2)} we are inspired by the patch matching and propose to discover discriminative patches from the image itself as well as the patch correspondences across images.

\subsection{Deep Attention Models}

In person re-ID, person misalignment and background biases \cite{Part-Bilinear} hinder the learning of a robust representation. Visual attention mechanisms aim at emphasizing the most informative regions for identification, while discarding the irrelevant ones (e.g., background and occluded regions). A binary hard attention was used in \cite{MSCAN} to localize the informative body parts of a person. Liu et al. \cite{CAN} proposed the Comparative Attention Network (CAN), which repeatedly localized the discriminative parts and compared the different regions of paired person images. Likewise, Wu et al. \cite{WU-Co-attention} introduced a deep co-attention based comparator to fuse co-dependent representations of paired images so as to correlate the best relevant parts. In Harmonious Attention Convolutional Neural Network (HA-CNN) \cite{HA-CNN}, a hard region-level attention and a soft pixel-level attention were learned in a unified attention block. Wang et al. \cite{Curriculum-sampling} considered both channel-wise and spatial-wise attention in a fully attentional block (FAB), where the channel information was re-calibrated and the spatial structure information was also preserved. Concurrently with attention models, bilinear pooling \cite{Compact-Bilinear} was first introduced to model the local pairwise feature interactions for fine-grained recognition problems. Thereafter, Wu et al. \cite{What-and-where} utilized an integrative form of the bilinear operation to pool a high-dimensional feature representation for the task of person re-ID. Recently, Suh et al. \cite{Part-Bilinear} used part-aligned representations to reduce the part misalignment by fusing the appearance and part feature maps in a bilinear pooling layer. Another state-of-the-art method was recently presented by Fang et al. \cite{BAT-net} to build a bilinear attention network by utilizing the higher-order statistical information that may hide in the feature maps. Unlike the above methods, we utilize the channel attention mechanism to detect body parts with detection parameters, which can be adaptive into different person shots.

%% file: method.tex
\begin{figure}[t]
  \includegraphics[width=\linewidth]{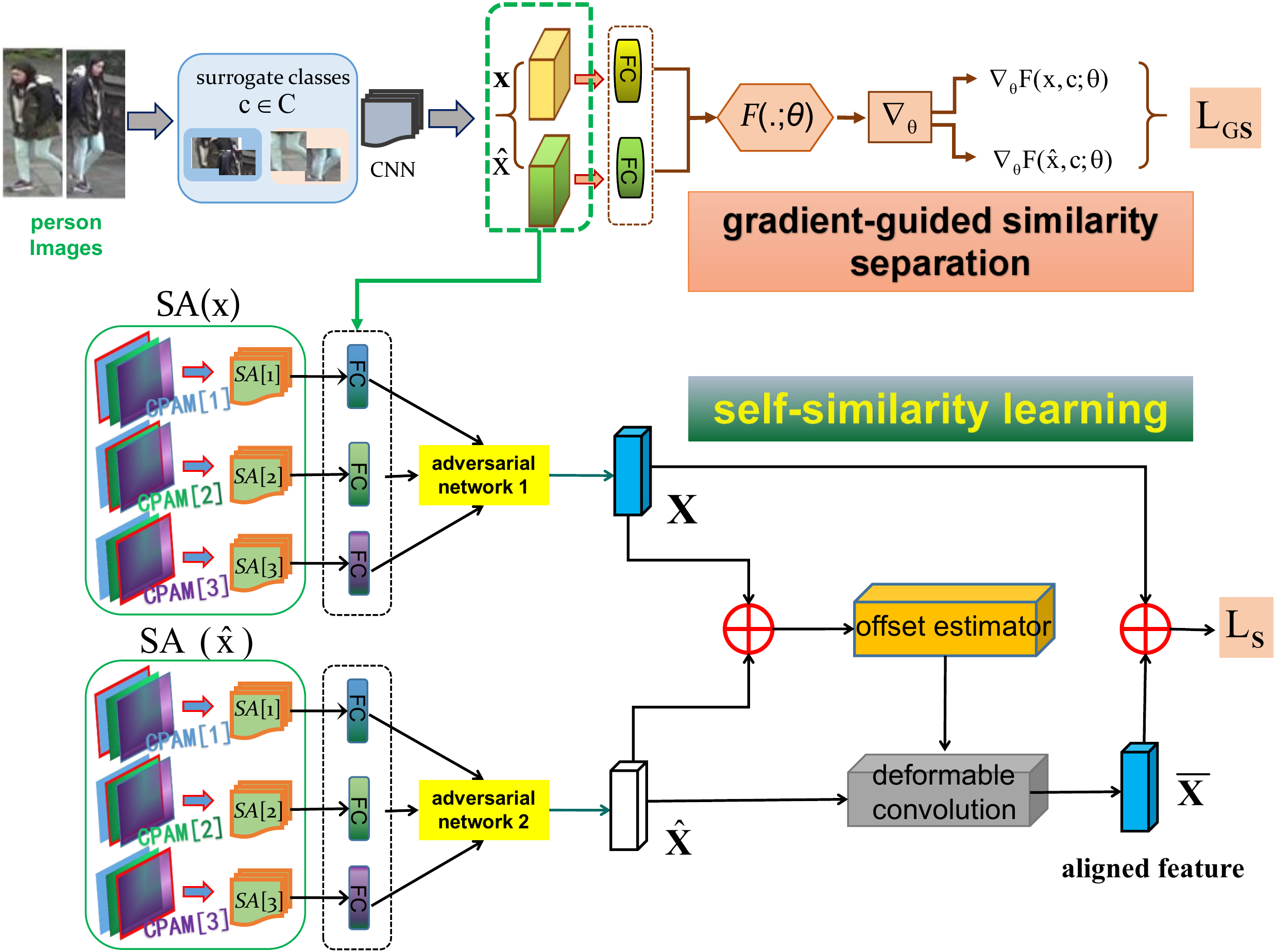}
  \caption{\small The proposed unsupervised self-similarity learning for person re-ID based on pseudo pairs. Given unlabelled person images, we first construct a set of surrogate classes for providing initial pseudo labels. Then we derive the gradient vectors of paired images $\mathbf x$ and $\mathbf{\Tilde{x}}$ w.r.t the surrogate class centroid vectors to form pseudo pairs. To improve the generalization, we propose the self-similarity learning, which consists of the intra-similarity (subject attention modules) and inter-similarity (deformable convolution with non-local blocks) learning.}\label{fig:framework}
\end{figure}

\section{Unsupervised Self-Similarity Learning for Person Re-identification}\label{sec:method}

Given unlabelled person images, we aim to jointly learn discriminative features and accurate similarity function in purely unsupervised setup. To acquire supervision information from unlabeled data, we construct the surrogate classes for each image via a group of transformations on its sampled patches (Section \ref{ssec:surrogate-class}). This can assign initial pseudo labels for images based on the global clustering. However, such a clustering criterion is noisy, and would impact the representation. Instead, we propose to cluster images into pseudo pairs by measuring the gradient alignment of pairwise images w.r.t the surrogate class. Then we propose a pairwise loss to reflect the clustering correctness (Section \ref{ssec:gradient-match}). To enhance the generalization of the learned similarity, we propose self-similarity learning that consists of intra-similarity learning to discover local discriminative features from individual images, and inter-similarity learning to align features across images for similarity comparison (Section \ref{ssec:self-similarity}). An overview of our framework is illustrated in Fig. \ref{fig:framework}. The notations used in this paper are summarized in Table \ref{tab:notations}.

\subsection{ Surrogate Class Construction}\label{ssec:surrogate-class}
We generate the surrogate classes by extracting patches sampled from each image, and then apply a range of transformations on these patches. The initial set of patches $\mathbb{P}=\{\mathbf{p}_1, \ldots, \mathbf{p}_N\}$ is formed by sampling $N$ patches of size $64\times32$ from different person images of size $128\times64$. For example, an image $I$ of size $128\times 64$ can be divided into four patches of size $64\times 32$. These patches can form the initial set $\mathbb{P}$ of $I$. Then, we define a family of transformations $\{T_{\gamma} | \gamma \in \Gamma\}$, parameterized by a set of vectors $\Gamma$. Each transformation is a composition of primitive transformations including flipping, translation, rotation and scaling. After each transformation, we apply color jittering on each patch. All primitive transformation parameters are concatenated into a single parameter vector $\gamma$.

\begin{table}[t]
\caption{A summary of notations.}
\begin{center}
\begin{tabular}{ |c|l| }
 \hline
 Notations & Descriptions\\
 \hline
 $\mathcal{L}_C(\cdot)$ & The loss for similarity learning\\
 $\mathcal{L}_{GS}(\cdot)$ & The gradient-guided similarity loss\\
 $\mathcal{L}_{Adv}(\cdot)$ & The adversarial loss for intra-similarity\\
 $\mathcal{L}_S(\cdot)$ & The multi-class cross-entropy loss for inter-similarity\\
 $\mathbf{x}$ &  The input image\\
 $\mathrm{C}$ & The surrogate classes \\
 $\theta$ & The model parameters \\
 $F(\cdot)$ & The similarity function\\
 $g_{\ast}$ & The gradient vector of the input image \\
 $\delta$ & The similarity threshold \\
 $A(\cdot)$ & The adversary network\\
 SA$[\cdot]$ & The subject attention learner\\
 CPAM$[\cdot]$ & The channel-attention module \\
 $h_k$ & The learnable offset for the $k$-th location \\
 $d_{st}$ & The similarity score between images $\mathbf{x}_s$ and $\mathbf{x}_t$\\
 $\psi$ & The regularization parameter for each SA$[\cdot]$\\
 \hline
\end{tabular}
\end{center}
\label{tab:notations}
\end{table}

\begin{figure}[t]
  \centering
  \includegraphics[width=0.8\linewidth]{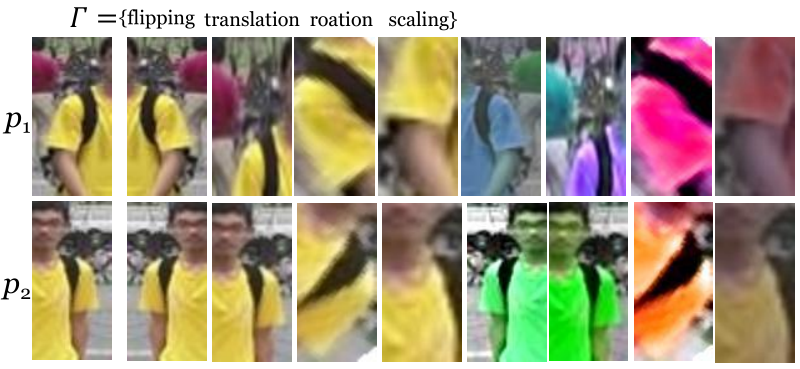}
  \caption{The surrogate class construction using patch transformations including flipping, translation, rotation and scaling. See texts for details. Color jitting is applied after each transformation.}\label{fig:patch-construct}
\end{figure}

As shown in Fig. \ref{fig:patch-construct}, for each patch $\textbf{p}_i$, we select $K$ parameter vectors $\{\gamma_i^1, \ldots, \gamma_i^K\}$ and apply the corresponding transformations $T_i=\{T_{\gamma_i^1},\ldots, T_{\gamma_i^K}\}$ to obtain a set of transformed patches $\Omega(\mathbf{p}_i)=T_i \mathbf{p}_i$, $i=1,\ldots,N$. For each patch set $\Omega(\mathbf{p}_i)$, we compute the RGB means at pixel-level across different patches to form a mean vector $\mathbf{m}_i$. Then, we perform $k$-mean clustering across these $\mathbf{m}_i (i=1,...,N)$ to form $\mathrm{C}$ clusters, for which we call \textit{surrogate classes}. For each cluster $c\in \mathrm{C}$, we have the centroid vector $\mathbf{v}_c$. A straight way of assigning a pseudo label to each image $\mathbf{x}$ is to use the 1-nearest neighbor across the surrogate classes, that is, directly compare $\mathbf{x}$ with $\mathbf{v}_c$ and choose the nearest class. However, such surrogate classes simply group visually similar patches while one identity image may be semantically associated with multiple classes. Thus, the nearest neighbor based labelling is noise-prone. In light of this, we propose to cluster images in pseudo pairs. Specifically, we define a similarity function $F(\mathbf{x},\mathbf{v}_c;\theta): \mathbf{x}\times \mathbf{v}_c \rightarrow \mathbb{R}$, which estimates the degree of confidence that the image $\mathbf{x}$ belongs to a surrogate class represented by $\mathbf{v}_c$. $F(\cdot)$ is implemented by a neural network parameterized by $\theta$. $F(\cdot)$ will be defined in Section \ref{ssec:gradient-match}. The surrogate classes simply provide initial supervision by evaluating the similarity value $F(\mathbf{x},\mathbf{v}_c;\theta)$ between $\mathbf{x}$ and $\mathbf{v}_c$ offline. Ideally, the classes should be defined at the identity-level wherein each image is related to a class with its pseudo label corresponding to the true identity. As such, each image is evaluated with the similarity between the image and the surrogate class. With this, in the following we propose a pairwise clustering criterion, which relates images into a class by measuring the gradient vectors between them.

\subsection{Gradient-Guided Similarity Separation}\label{ssec:gradient-match}

To assign pseudo labels to unlabelled person images, we propose to measure the gradient vectors of pairwise images w.r.t the surrogate classes. We observe that when the similarity function $F(\cdot)$ learns to measure the similarity of two images from the same class manifold \footnote{Ideally, each class manifold corresponds to an identity.}, the gradient vectors of a loss function defined over $F(\cdot)$ at the two images are highly correlated with each other. Thus, $F(\cdot)$ can be formulated to maximize the correlation between gradients over the pairwise images if the images are from the same identity. In this spirit, we propose a gradient-guided similarity separation, which drives the gradient alignment over pairwise images. 

\begin{figure}[t]
  \centering
  \includegraphics[width=0.8\linewidth]{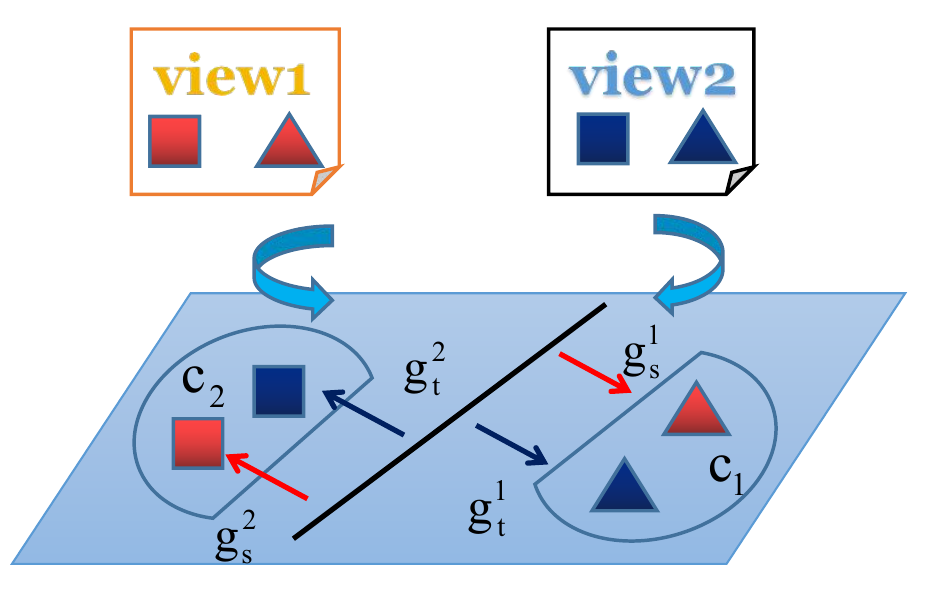}
  \caption{Scheme of gradient-guided similarity separation for improved pseudo labelling. The same shape indicates samples from the same identity class. Given two classes $c_1, c_2 \in C$: the gradient vectors of pairwise samples $\mathbf{g}_s^1,\mathbf{g}_t^1$ and $\mathbf{g}_s^2,\mathbf{g}_t^2$ are encouraged to be centered around the corresponding class by minimizing the gradient-guided similarity loss for $d_{st}^1$ and $d_{st}^2$, respectively. Best viewed in color.}\label{fig:GS-class}
\end{figure}

Formally, given two person image features $\mathbf{x}$ and $\mathbf{\Tilde{x}}$ across camera views, we define $\mathbf{g}_s$ and $\mathbf{g}_t$ as the respective gradient vectors derived from a loss function for $F(\cdot)$. Thus, we have
\begin{equation}
\begin{split}
    \mathbf{g}_s = \nabla_{\theta} \mathcal{L}_C (F,\mathbf{x},\mathbf{v}_c;\theta), \mathbf{g}_t = \nabla_{\theta} \mathcal{L}_C (F,\mathbf{\Tilde{x}},\mathbf{v}_c;\theta).
\end{split}
\end{equation}
Here, $\mathcal{L}_{C} (F, \ast, \mathbf{v}_c;\theta)$ denotes the soft-max function for optimizing the similarity function $F(\ast,\mathbf{v}_c;\theta)$, where $\mathbf{v}_c$ is a specific surrogate class vector. The similarity function between the image feature $\mathbf{x}$ and the class vector $\mathbf{v}_c$ takes the form as:
\begin{equation}
    F(\mathbf{x},\mathbf{v}_c;\mathbf{W},b)= \mathbf{x}^T \mathbf{W} \mathbf{v}_c+b,
\end{equation}
where $\mathbf{W}$ is the computational matrix corresponding to the parameter $\theta$, and $b$ is a bias (We omit $b$ in the rest of the paper for simplicity). As the gradient vectors can characterize the direction towards the local minima \cite{WU-Video-Re-ID}, we propose to measure the discrepancy between $\mathbf{g}_s$ and $\mathbf{g}_t$ via the cosine similarity. Accordingly, the gradient-guided similarity, i.e., $d_{st}$, can be formulated as the cosine distance between $\mathbf{g}_s$ and $\mathbf{g}_t$:
\begin{equation}
    d_{st}=\frac{\mathbf{g}_s^T \mathbf{g}_t}{||\mathbf{g}_s||_2 ||\mathbf{g}_t||_2}.
\end{equation}
Intuitively, minimizing $d_{st}$ amounts to encouraging \textit{pairwise} images from the same identity to center around the same cluster. To update the pseudo labeling during training, we introduce a new loss $\mathcal{L}_{GS}$ based on the pairwise alignment between gradients. This is inspired by the manifold learning indicating that complex low-dimensional manifolds can be learned from pairwise distance optimization \cite{Unsupervised-deep-cluster}. Person re-ID can be interpreted as manifold clustering: if all intra-personal images are closer in the feature space than the inter-personal images, then a perfect person retrieval is achieved \cite{GLAD,E-Metric,SSM,LDA-Re-ID}. Based on the above, the gradient-guided similarity separation is formulated by minimizing the following loss:
\begin{equation}\label{eq:GS-loss}
    \mathcal{L}_{GS}=\mathbb{E} \left[1-\frac{\alpha}{2}(d_{st}-\delta)^2 \right],
\end{equation}
where $\delta\in (0,1)$ refers to a similarity threshold and $\alpha>0$ controls the slope of the gradient direction. In each training batch, we approximate the $\mathcal{L}_{GS}$ by computing the expectation of parameters sampled from the training iterations. We remark that we do not constrain the value range of the similarity scores $d_{st}$ by re-scaling because $d_{st}$ has been normalized. The rationale of Eq.\eqref{eq:GS-loss} is to implicitly reveal the identity manifold across images by promoting the \textit{pairwise} gradient distances between images. We operate purely on image pairs without global clustering to maximize the pairwise similarity for images from the same identity. The idea of gradient-guided similarity separation is depicted in Fig. \ref{fig:GS-class}.

During training, our methods updates the cluster centroid vector $\mathbf{v}_c$ via center-based aggregation. Suppose that there are $K$ clusters with the $k$-th cluster having $N_k$ images. Let $f_{y_i}(\mathbf{x})$ be the neural network penultimate layer, which outputs the image $\mathbf{x}$ with the pseudo label $y_i$. The center vector $\mathbf{v}_c$ is then computed as $\mathbf{v}_c=\frac{1}{N_k}\sum_{y_i=k}f_{y_i}(\mathbf{x})$. Our objective is to jointly learn the pseudo labelling (with $\mathcal{L}_{GS}$) and the similarity function $F(\cdot)$ (with $\mathcal{L}_C$) via the gradient descent over all training images in pairs. Thus, we formulate the learning objective function as:
\begin{equation}\label{eq:obj-cluster-sim}
    \mathbb{L}=\mathcal{L}_C + \mathcal{L}_{GS}.
\end{equation}
In Eq. \eqref{eq:obj-cluster-sim}, when optimizing $\mathcal{L}_{GS}$ by a batch gradient descent update rule, one needs to compute $\mathbf{g}_s$ and $\mathbf{g}_t$ over paired images. Although the initial surrogate clustering may assign different (incorrect) labels to the images, the gradient alignment scheme can effectively associate images into the correct class. Meanwhile, the similarity function $F(\cdot)$ can also be jointly optimized with the improved clustering.

\subsection{Fine-Grained Self-Similarity Learning}\label{ssec:self-similarity}

The above gradient-guided similarity alignment is able to cluster images based on pairwise comparison. However, optimizing the similarity learning supervised by identity-level manifolds is limited in generalizing the learnt metric to \textit{unseen} shots. This is because the identity-level semantics may easily lead the training model to select the easiest examples so as to decrease the overall empirical training risk. This tends to make the training model over-fitting to training samples. To generalize the re-ID model, recent studies attempt to learn part-based features \cite{PAUL,Dual-part-align,Spindle-net}, which are discriminative and generalizable to unseen identities \cite{DeML}. Inspired by this, we propose a \textit{fine-grained self-similarity} to improve the generalization of the similarity function through 1) intra-similarity learning on each image to discover diverse part features, and 2) inter-similarity learning across images to align features under visual variations. Unlike existing patch sampling based re-ID methods \cite{PAUL,Dual-part-align,Spindle-net} that require precise bounding-box annotations, our method requires no auxiliary annotations but \textit{learns} to attend part features within/across label-free images.

\subsubsection{Intra-Similarity Learning via Adversarial Learning}

To identify the discriminative patterns on each image, we utilize the cues of the image itself. It is observed that convolutional filters can be used as attribute detectors, and the resulting feature maps preserve diverse attributes in the channels \cite{DeML}. Inspired by this, we present a \textit{Channel Part Attention Module} (CPAM) to detect the part features in CNN channels (e.g., hoodie, striped T-shirt, backpacks). This strategy leads to the intra-similarity learning. 

\begin{figure}[t]
  \centering
  \includegraphics[width=\linewidth]{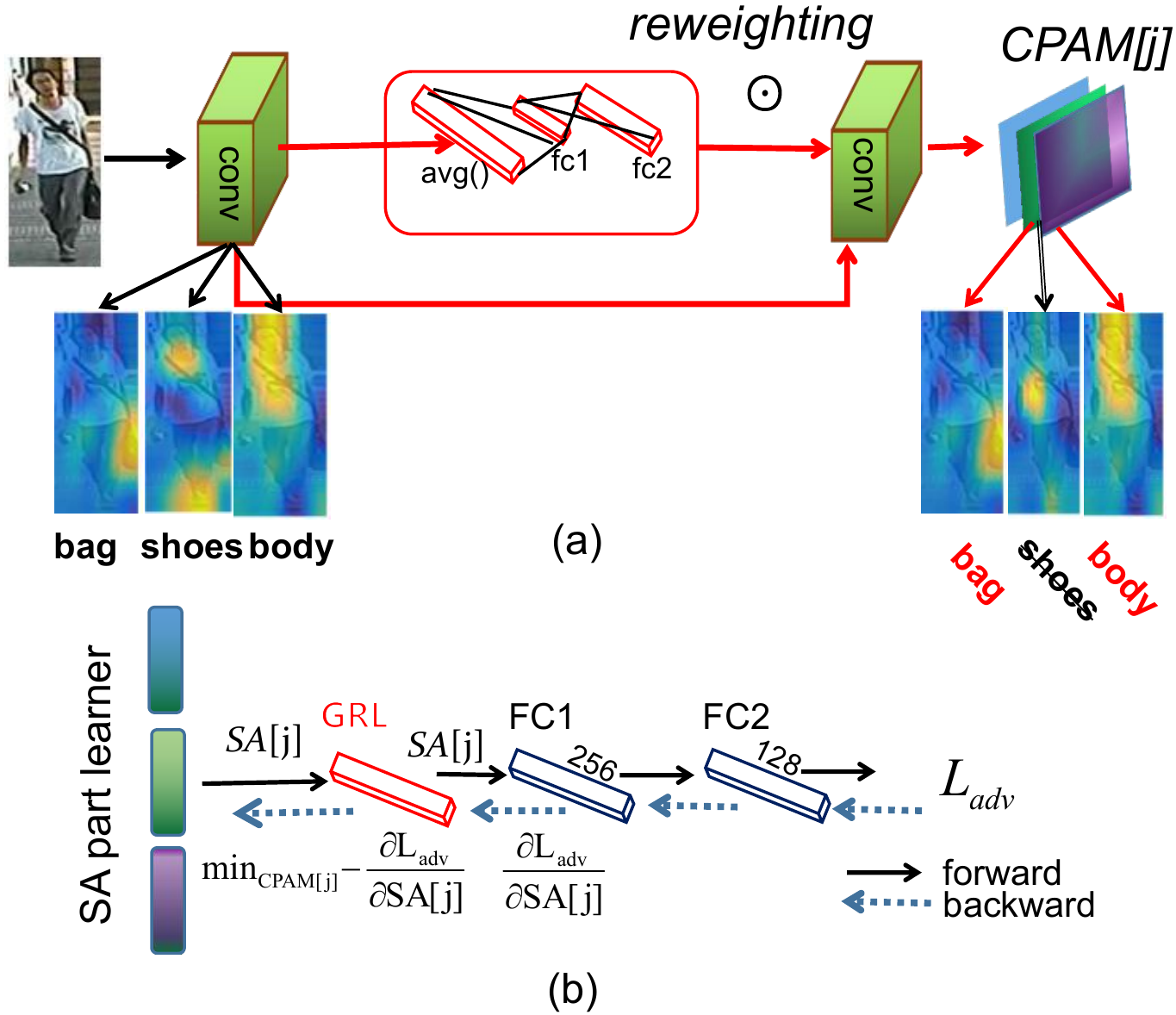}
  \caption{Scheme of the intra-similarity module for each identity. (a) The computation of the CPAM$[j]$ via subject attention. In this case, ``bag" and ``body" are activated while ``shoes" is suppressed. (b) The adversary network is placed after each subject-attention learner. (Best viewed in color). }\label{fig:intra-similarity}
\end{figure}

Suppose there are $J$ attributes to be detected, we can detect each attribute in CPAM$[j],j=1,\ldots,J$. Let $U\in \mathbb{R}^{C\times H\times W}=f(\mathbf{x})$ denote the feature block truncated from the deep neural network, e.g., ResNet \cite{Resnet}. Here $\mathbf{x}$ is the image. $f(\cdot)$ denotes the network parameters. $C$, $H$ and $W$ denote the number of channels, height and width of the feature block. For each CPAM$[j]$, we learn its corresponding Subject (identity) \footnote{We use subject and identity interchangeably.} Attention learner, i.e., SA$[j], (j=1,\ldots,J)$ to capture its  attribute. SA$[j]$ will be defined later. Specifically, the CPAM$[j]$ is computed as follows:
\begin{equation}
    \mbox{CPAM}[j]=U \odot \sigma[ \mathbf{W}^2_j ReLU [ \mathbf{W}^1_j \Psi(U)]],
\end{equation}
where $\Psi(\cdot)$ is a spatial average pooling to aggregate $U$ into a channel descriptor, which is then passed through two fully-connected layers, parameterized by $\mathbf{W}^1_j\in \mathbb{R}^{64\times C}$ and $\mathbf{W}^2_j\in \mathbb{R}^{C\times 64}$ to capture the interactions between channels. The resulting feature block is then operated by ReLU and Sigmoid $(\sigma[\cdot])$. After that, $U$ is reweighted by the channel-wise multiplication $\odot$ between $U_{\bar{c}}$ and the scalar $\sigma[ \mathbf{W}^2 ReLU[\mathbf{W}^1 \Psi(U)]]_{\bar{c}}$, where $\bar{c}$ denotes the channel index. The computation of CPAM $[j]$ is depicted in Fig. \ref{fig:intra-similarity} (a).

\paragraph{Understanding the Diverse Attribute Detector}
The re-weighting operation in the CPAM$[j]$ can be understood as a group of attribute selectors (e.g., in Fig.\ref{fig:intra-similarity}, it attends to ``jacket" and ``pants", yet ignores ``head"). The region proposals generated from each CPAM$[j]$ can be combined into a detector for different attributes. However, such a combination has to determine which channel contains what attribute. This turns out to be very difficult. To address this challenge, we feed each CPAM$[j]$ into a stack of \textit{shared} embedding layers (parameterized by $g(\cdot)$) so as to transfer the diversity of different CPAM$[j]$ into a combined learner, i.e.,  SA$[j]=g(\hat{U})$, where $\hat{U}$=CPAM$[j]$. Then, we explicitly encourage the diversity among these CPAM$[j],j=1,\ldots,J$ by imposing a diversity constraint on the SA$[j]$ via an adversary network. Such an adversary network tries to minimize the discrepancies amongst the SA learners while the CPAM tries to simultaneously detect different attributes of different channels to maximize these discrepancies. Hence, the adversarial optimization can be cast as a max-min game:
\begin{equation}\label{eq:subject-adv}\small
 \max_{f, g, \hat{U}} \min_A \mathcal{L}_{Adv} =\sum_{j,j'}^J || A(\mbox{SA}[j] (f(\mathbf{x}))) - A(\mbox{SA}[j'] (f(\mathbf{x}))) ||_2^2,
\end{equation}
where $\mathbf{x}$ is the input image. $f$ and $g$ denote the mapping functions of the convolutional blocks. $A(\cdot)$ is the adversary network. Eq. \eqref{eq:subject-adv} measures the discrepancy across all SA learners of one identity. To simplify the two-stage optimization, we adopt the gradients reverse layer (GRL) \cite{GRL} to make the objective equivalent to $\min_{A} \mathcal{L}_{Adv}$ and $\min_{f,g,\hat{U}}(-\mathcal{L}_{Adv})$.

The above part feature detector should be adaptive to different images from the same identity. Thus, we apply a regularization on each SA to efficiently fine-tune the new parameters. The SA learners can learn to detect the attributes with the parameters $\boldsymbol \psi^*$ via $\boldsymbol \phi_i = \mathcal{L}_{Adv} \left(\arg\min \mathcal{L}_{Adv} (\boldsymbol \psi^*, \mathbf{x}_i), \mathbf{x}_j \right)$. The learned parameters $\boldsymbol \psi^*$ should have a good generalization on the new sample $\mathbf{x}_j$, which has an inductive bias particularly well-suited for different person shots. To this end, we consider an explicit regularized term $ ||\boldsymbol \phi -\boldsymbol\psi ||^2$, which can be added into Eq. \eqref{eq:subject-adv}. Thus, we have
\begin{equation}\label{eq:regularizer}\small
\arg\min_{\boldsymbol\phi} \mathcal{L}_{Adv} (\boldsymbol\phi, x_i) + \frac{\beta}{2}  || \boldsymbol\phi -\boldsymbol\psi ||^2,
\end{equation}
where the regularization term $\frac{\beta}{2} || \boldsymbol\phi -\boldsymbol\psi ||^2$ encourages $\boldsymbol\phi$ to remain close to $\boldsymbol\psi$, while $\beta$ plays a role in controlling the strength of prior $\boldsymbol \psi$ relative to the new attribute. By retaining on the prior $\boldsymbol \psi$, the parameters can be reused and the computational burden is reduced.

In summary, the intra-similarity learning mechanism performs self-similarity learning by introspecting each person image into detectable attributes, and thus distilling discriminative parts for accurate similarity estimation. However, the intra-similarity learning performed on individual images remains not optimal in calculating the disparity across images caused by viewpoint changes. In the following, we propose the inter-similarity learning to compare local features across images.

\subsubsection{Inter-Similarity Learning based on Deformable Convolutions}

\begin{figure}[t]
  \centering
  \includegraphics[width=\linewidth]{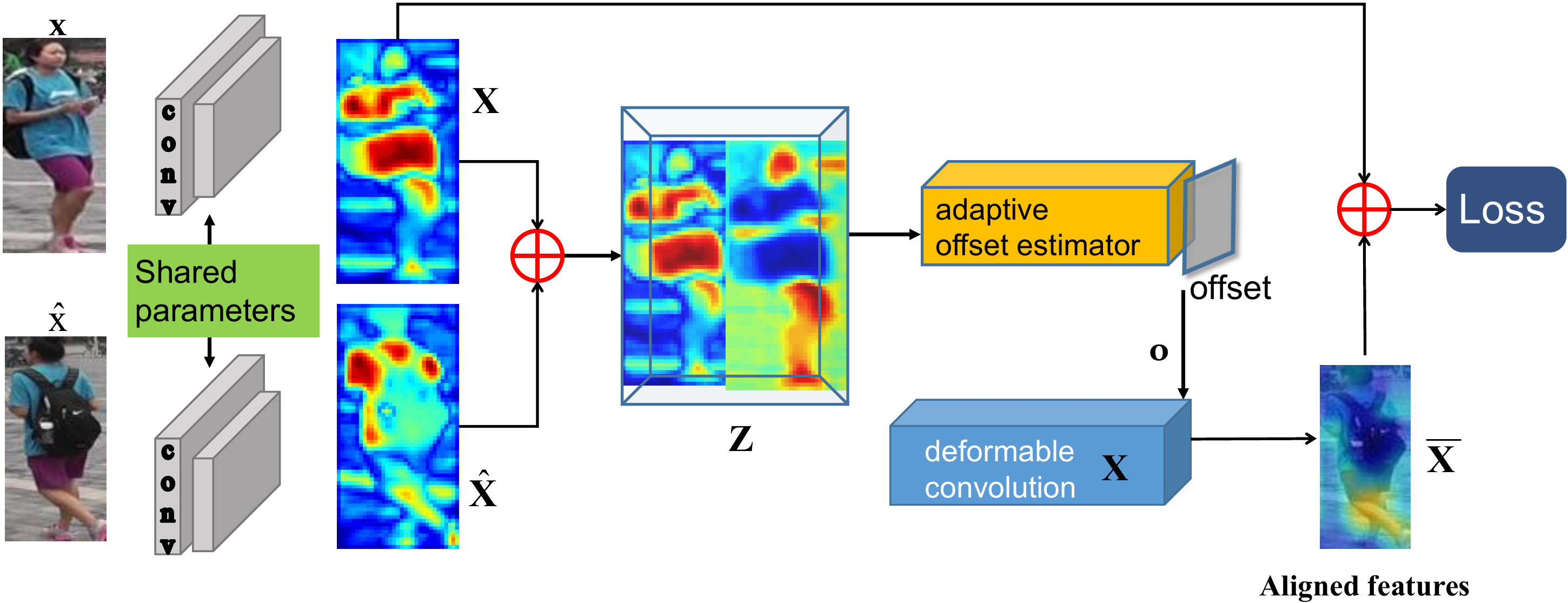}\\
  (a)\\
  \includegraphics[width=\linewidth,height=6cm]{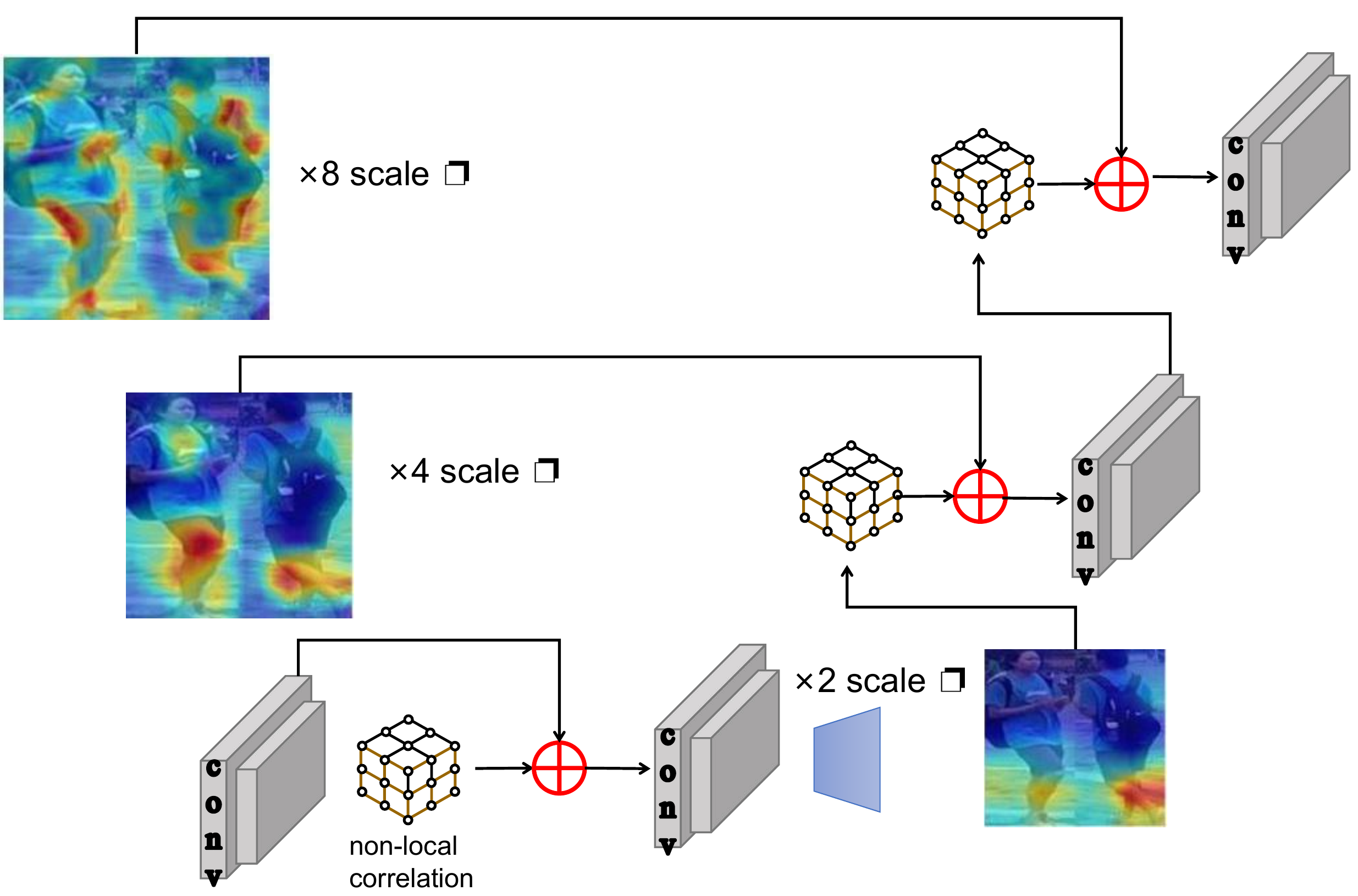}\\
  (b)
  \caption{The inter-similarity module for cross-image feature alignment. (a) The inter-similarity operation uses deformable convolution to align features across views. (b) Inside the offset estimator, we adopt the non-local block connection with multi-scale features to improve the alignment. (Best viewed in color).}\label{fig:inter-similarity}
\end{figure}

The inter-similarity learning aims to estimate the local similarity across person images that may undergo visual variations caused by viewpoint changes. Taking the advantage of deformable convolution in local feature alignment \cite{T-DAN,SSEN}, we formulate the inter-similarity learning as an \textit{integrative} process of matching the patches between an image and its reference image \footnote{A reference image is an image selected from a different camera view.}. As such, we aim to search the cross-view correspondences for similarity estimation. Given the input-reference pair ($\mathbf{X},\hat{\mathbf{X}}$), we extract features from $\hat{\mathbf{X}}$ in an aligned fashion, i.e., matching the patches w.r.t $\mathbf{X}$ by estimating the \textit{cross-image offset}. Specifically, we adopt the deformable convolution with offset to search for the patch correspondences. However, estimating the offset for deformable convolution is difficult because the patches may exhibit spatial variations caused by viewpoint changes. To address this, we design non-local blocks in the deformable convolution to automatically estimate such offset. As such, the inter-similarity is implemented by learning to align cross-view patches through the deformable convolution. The idea of inter-similarity learning is shown in Fig. \ref{fig:inter-similarity}.

\paragraph{Deformable Convolution with Adaptive Offset}

The deformable convolution was proposed to improve the CNN in permitting geometric transformations \cite{Deform-Conv}. It is characterized by a learnable offset to facilitate the sampling of pixel points with deformed grids. By virtue of this, we deploy the deformable convolution to align cross-image patches for both feature extraction and similarity estimation. To estimate the offset for deformable convolution, non-local blocks based on a multi-scale structure are adopted. 

Let $\mathbf{X}$ and $\mathbf{\hat{X}}$ be the input and reference feature block, the deformable convolution can be defined as:
\begin{equation}\label{eq:deformable}
\mathbf{\Bar{X}}[j]= \sum_{k=1}^K \mathbf{w}_k \cdot \mathbf{X}[j + j_k + \bigtriangleup o_k],
\end{equation}
where $\mathbf{w}_k$, $j$, $j_k$ and $\bigtriangleup o_k$ denote the $k$-th kernel weight, indices of the center, the $k$-th fixed offset, and the learnable offsets for the $k$-th location. Eq.\eqref{eq:deformable} aligns the reference feature $\mathbf{\hat{X}}$ w.r.t $\mathbf{X}$ via the deformable convolution by adjusting the offset. During the alignment, the offset $\bigtriangleup o_k$ should be adaptive to the input $\mathbf{X}$ rather than a fixed value. To this end, in the following we present a novel offset estimator based on non-local feature connection with multi-scale features.

\paragraph{Non-Local Feature Block Connection with Multi-Scaling} To learn an adaptive offset for the deformable convolution, we design an \textit{adaptive offset estimator}. As illustrated in Fig.\ref{fig:inter-similarity} (a), $\mathbf{X}$ and $\mathbf{\hat{X}}$ are concatenated into the feature block $\mathbf{Z}$, which is input to the adaptive offset estimator. Inside the offset estimator, we employ a non-local block \cite{Non-local,SSEN} to localize the relevant features that are spatially distant. The non-local connection captures the high-order correlations between inter-features, and thus can predict the dynamic offset. To robustly deal with large visual displacements in cross-view person images, we incorporate three non-local blocks into the adaptive offset estimator where the local features are amplified at various scales (i.e., $2\times$, $4\times$ and $8\times$ scales, see Fig. \ref{fig:inter-similarity} (b)). Specifically, the non-local block connection is defined as:
\begin{equation}\label{eq:non-local}
\mathbf{o}_j = \mathbf{Z}_j + \mathbf{W}_{\mathbf{o} } \frac{1}{\mathrm{N}(\mathbf Z)} \sum_i u(\mathbf{Z}_j, \mathbf{Z}_i) \mathbf W_v \mathbf{Z}_i,
\end{equation}
where $\mathbf{o}$ is the offset. $i$ is the index of all possible positions and $j$ is the index of the output position. $\mathbf{W}_{\mathbf{o}}$ denotes the weight matrix and $\mathrm{N}(\mathbf Z)$ is the normalization factor. $u(\cdot)$ represents the pair-wise computation and $\mathbf W_v \mathbf Z_i$ computes the linear embedding of the input $\mathbf Z$ at position $i$. In Eq.\eqref{eq:non-local}, the learned offset $\mathbf{o}$ encodes the high-order correlation between $\mathbf{X}$ and $\mathbf{\hat{X}}$ at patch-level. Further, $u(\mathbf{Z}_j, \mathbf{Z}_i)$ calculates the pairwise similarity which can be defined as a Gaussian function $u(\mathbf{Z}_j, \mathbf{Z}_i)= \exp (\sigma_1(\mathbf{Z}_j)^T \sigma_2(\mathbf{Z}_i) )$ where $\sigma_1 (\cdot)$ and $\sigma_2 (\cdot)$ are linear embedding functions. Hence, the non-local block computed on the concatenated feature block can be considered as measuring the patch-wise similarity across images. Finally, the network maps the feature $\widetilde{\mathbf X}_i =\mathbf X_i \oplus \mathbf{\Bar{X}}_i$ to a classification score vector $z_i = \Theta (\widetilde{\mathbf X}_i)$, which is then normalized by a softmax function to produce a probability distribution $p(y_{i l_j} | \widetilde{\mathbf X}_i )=\frac{\exp(z_{ij})}{\sum_{j=1}^C \exp(z_{ij})}$. We have the loss function $\mathcal{L}_S$ defined as
\begin{equation}\label{eq:multi-class-CE}
    \mathcal{L}_S =\log p( y_{i l_i}| \widetilde{\mathbf X}_i=\mathbf X_i \oplus \mathbf{\Bar{X}}_i ),
\end{equation}
where $\oplus$ denotes the concatenation. Herein, the loss is computed as the multi-class cross-entropy between the true label $l_i$ and the predicted probability for an estimated label $y_{i l_j}$.

\subsubsection{The Objective Function} 
Combining the similarity function loss ($\mathcal{L}_C$), the gradient-guided similarity separation loss (Eq. \eqref{eq:GS-loss}), the self-similarity learning loss (Eq. \eqref{eq:subject-adv} and Eq. \eqref{eq:multi-class-CE}), the final objective function of our training model can be formulated as:
\begin{equation}\label{eq:obj-function}
    \mathbb{L}= \mathcal{L}_C + \mathcal{L}_{GS} + \lambda_1 \mathcal{L}_{Adv} - \lambda_2  \mathcal{L}_S,
\end{equation}
where $\lambda_1$ and $\lambda_2$ are trade-off hyper-parameters.

%% file: Experiment.tex
\section{Experiments}\label{sec:exp}

In this section, we present experimental evaluations of the proposed approach on three benchmark datasets: \textbf{Market-1501} \cite{Market-1501}, \textbf{CUHK03-NP} \cite{k-reciprocal,Deep-re-ID} and \textbf{MSMT17} \cite{PTGAN}. First, we describe the experimental setup, and then evaluate a few variants of our method by comparing with the state-of-the-art (SOTA) methods. We thorougly analyze our model via ablation studies, which include the influence of surrogate classes and the effect of gradient-guided separation and the self-similarity learning component.

\subsection{Datasets and Evaluation Protocol}


\begin{itemize}
    \item \textbf{Market-1501:} This dataset consists of 32,668 person images of 1,501 identities observed under six different camera views. The dataset is split into 12,936 training images of 751 identities and 19,732 testing images of the remaining 750 identities. Both training and testing images are detected using a DPM detector \cite{DPM}.
    \item \textbf{CUHK03-NP:} This dataset has a new protocol on training and testing partition on CUHK03 dataset \cite{k-reciprocal,Deep-re-ID} with 767 and 700 identities for training and test, respectively. The CUHK03 dataset provides two types of data with manual labels and DPM detection bounding boxes. We conduct experiments on both types of data.
    \item \textbf{MSMT17:} This is the largest person re-ID dataset \cite{PTGAN}, containing 126,441 person images from 4,101 identities. The person images are detected by Faster R-CNN \cite{Faster-r-cnn}. This dataset is collected under 15 different cameras. The training set consists of 32,621 images belonging to 1,041 identities, whereas the test set contains 93,820 images of 3,060 identities.The test set is further randomly split into 11,659 query images and 82,161 gallery images.
\end{itemize}

We adopt a widely-used evaluation protocol \cite{Deep-re-ID,Improved-re-ID}. In the matching process, we calculate the similarities between each query and all the gallery images, and then return the corresponding ranked list. All the experiments are conducted under a single query setting. The performances are evaluated using the cumulative matching characteristics (CMC) curves, which are an estimate of the expectation of finding the correct match in the top $k$ matches. We also report the mean average precision (mAP) scores \cite{Market-1501} over the above datasets.

\subsection{Implementation Details}

Our implementations are based on the PyTorch framework \cite{Pytorch}. We used the codes released by \cite{DeML,Adv-ML}, and we considered two backbone networks: ResNet-50 \cite{Resnet} and the pre-trained GoogLeNet V1 \cite{Google-net}. ResNet-50 is widely used as the backbone in person re-ID methods. We also implemented our model based on GoogLeNet-V1 because the inception module of GoogLeNet uses a range of convolutional filters ($1\times 1, 3\times 3$ and $5\times 5$) together with $3\times3$ max pooling performed in a \textit{parallel} way. As such, the learned features can capture objects at multiple scales. This design has demonstrated to be beneficial to part-based person re-ID \cite{Part-Bilinear} and attribute learning for re-ID \cite{ACRN} where GoogLeNet is used to align body parts (or detect person attributes) caused by varying-sized objects.

In the set of transformations $T_{\gamma}$, the translation is performed vertically and horizontally by a distance within 0.2 of the patch size. The scaling is done by multiplication of the patch scale by a factor in the range [0.7, 1.4] and the rotation is to rotate the image by an angle of up to 20 degrees. We resize all training images to $128 \times 256$ and then augment them by horizontal flipping and random erasing \cite{Random-Erase}. The SA learners, CPAMs and adversary networks are initialized using random weights. We set the batch size to 64 and train the model with a base learning rate that starts from 0.05 and decays to 0.005 after 40 epochs, with training completed after 300 epochs. We set the moment $\mu$= 0.9 and the weight decay as 0.0005. In addition, we set $\lambda_1 = 0.015$, $\lambda_2 = 2.5$. The final feature dimensionality is set to $d=512$. To train the network with the three losses, we adopt a stage-wise scheme: we first pre-train the network using only the separation loss ($\mathcal{L}_{GS}$), then fine-tune it with the $\mathcal{L}_{adv}$ loss integrated into the training objective. While optimizing the objective function, we also update the classification model $\mathcal{L}_S$ by regularly re-initializing the classification parameters every $N$ iterations. During the computation of $\mathcal{L}_{GS}$, it is empirically found that minimizing the $l_2$ loss between $g_s$ and $g_t$, i.e., $||g_s-g_t||_2^2$, and the weight constraint on the attention learners $||SA[j](x)||_2^2$ leads to better results. All of the experiments were conducted on a single NVIDIA TITAN XP GPU \footnote{Codes will be available on Github once paper is accepted for publication.}.

\subsection{Comparison with State-of-the-Art Methods}

We empirically verify the effectiveness of our approach by comparing with a number of SOTA methods on three benchmark datasets. 

\paragraph{Market-1501} Comparison results on Market-1501 are reported in Table \ref{tab:Market-1501-CMC}. We consider the following methods: \textbf{1)} the pseudo label learning based methods: CASCL \cite{CASCL}, CAMEL \cite{CAMEL}, BUC \cite{BUC}, MAR \cite{MAR}, MMCL \cite{MMCL} and SSL \cite{SSL}; \textbf{2)} the UDA based methods: UST \cite{Unsupervised-TIP20}, TJ-AIDL \cite{TJ-AIDL}, UMDL \cite{UMDL}, Exemplar \cite{Exemplar}, SPGAN \cite{SPGAN}, PTGAN \cite{PTGAN}, Self-paced-CL \cite{Self-paced-CL}, MMT \cite{MMT} and HHL \cite{HHL}; \textbf{3)} the self-similarity learning methods: PT-RDC \cite{Radial-Distance-U-Re-ID} and SSG \cite{SSG}.

We have the following observations. The proposed method is seen to outperform all SOTA methods, especially in comparing with the recent self-similarity learning methods, i.e., PT-RDC \cite{Radial-Distance-U-Re-ID} and SSG \cite{SSG}. For instance, PT-RDC \cite{Radial-Distance-U-Re-ID} uses an expertly-designed pose transformation dataset to generate more images and improves the discriminative clustering. Comparing with PT-RDC \cite{Radial-Distance-U-Re-ID}, the proposed method U-SSL* (backboned on ResNet-50 with dense feature concatenation) does not require any augmentation and achieves a new SOTA performance of rank-1= 94.1\% and mAP=82.3\%, respectively. This is implemented by simply concatenating the global pooled features from each residule block to form the final features. Comparing with the SOTA UDA based methods, i.e, Self-paced-CL \cite{Self-paced-CL} and MMT \cite{MMT}, the proposed method also achieves superior results on both rank-1 and mAP. These UDA methods rely on the choice of souce domains, which may impact the performance of different target datasets. In comparison with UDA based models, our method U-SSL avoids the fragility of choosing a suitable source domain.

In addition, we also have the following observations. First, our method is compatible with different backbones, i.e., GoogLeNet \cite{Google-net} and ResNet-50 \cite{Resnet}. Second, the general heuristic based pseudo label generation (e.g., using clustering) is limited in discovering discriminative features. In contrast, we construct surrogate classes containing a variety of feature transformations, which can potentially learn optimal discriminative features.

\paragraph{CUHK03-NP}
We also evaluated the proposed method on CUHK03-NP and reported the comparison results with SOTA methods in Table \ref{tab:CUHK03-reID-CMC}. We compared with the following baselines: CAMEL \cite{CAMEL}, PTGAN \cite{PTGAN}, MLFN \cite{MLFN}, DaRe \cite{DaRe} and DG-Net \cite{DG-Net}. Table \ref{tab:CUHK03-reID-CMC} shows that our method outperforms all competitors consistently across all measures. A primary reason is that our method learns discriminative features from pseudo-pairs, with the learning jointly performed with the similarity learning. This leads to a more accurate similarity computation. More specifically, by exploiting the channel attention for detecting body parts, our method discovers the diverse parts from the different channels. This advantage is demonstrated in the detected CUHK03-NP dataset, where person images exhibit severe occlusion and background clutter, due to the detection imperfection. For example, comparing with the SOTA method DaRe \cite{DaRe}, our method improves the rank-1 and mAP by 5.0\% and 8.6\%, respectively.

\paragraph{MSMT17} Table \ref{tab:MSMT17-CMC} shows the comparison results on this new challenging dataset. As observed, our proposed method outperforms the baseline algorithms by a noticeable margin. More specifically, U-SSL based on GoogLeNet outperforms the best UDA based competitor MMT (Duke-to-MSMT) \cite{MMT} by 4.6\% in terms of the rank-1 value. It is noted that MMT \cite{MMT} needs to choose an appropriate source domain. For example, when MMT \cite{MMT} chooses Market-1501 as the source domain, its rank-1 value drops by 6.3\%, compared with the performance obtained by using DukeMTMC-reID as the source domain. When we use ResNet-50 as the backbone, the proposed U-SSL achieves 63.1\% at rank-1. When U-SSL densely connects features from GoogLeNet, it further improves the rank-1 to reach a new SOTA performance. This demonstrates the generalization of our method in leveraging diverse features at different layers. In addition, the intra-similarity based on attention channels can locate the discriminative body parts which are useful to distinguish identities.

\begin{table}[t]
  \caption{Comparison with SOTA on the Market-1501 dataset. U-SSL (*) means that we densely concatenate the features from each feature block. The best results are in boldface.}
  \label{tab:Market-1501-CMC}
  \center
  \begin{tabular}{|r|ccc|c|}
  \hline
    Model &  Rank-1 & Rank-5 & Rank-10 & mAP\\
    \hline
    CASCL \cite{CASCL}  & 65.4 & 80.6 & 86.2 & 35.5\\
    CAMEL \cite{CAMEL}  & 54.5 & -&- & 26.3\\
    UMDL \cite{UMDL}  & 34.5 & 52.6 & 59.6 & 12.4\\
    MAR \cite{MAR}  & 67.7 & 81.9 &- & 40.0\\
    BUC \cite{BUC}  & 61.9 & 73.5 & 78.2 & 29.6\\
    TJ-AIDL \cite{TJ-AIDL}  & 58.2 &-&- & 26.5\\
    SSL \cite{SSL}  & 71.7 & 83.8 & 87.4 & 37.8\\
    UST \cite{Unsupervised-TIP20}  & 73.7 & 84.0 & 87.9 &38.0\\
    Exemplar \cite{Exemplar}  & 75.1 & 87.6 & 91.6 & 43.0\\
    SPGAN \cite{SPGAN}  & 51.5& 70.1& - & 27.1\\
    PTGAN \cite{PTGAN}  & 38.6 & 57.1 &- & 15.7\\
    HHL \cite{HHL}  & 62.2 & 78.8 &- & 31.4\\
    MMCL \cite{MMCL}  & 80.3 & 89.4& 92.3 & 45.5\\
    Self-paced-CL \cite{Self-paced-CL}  & 88.1 & 95.1 & 97.0  & 73.1\\
    MMT \cite{MMT} & 87.7 & 94.9 & 96.9 & 71.2 \\
    PT-RDC \cite{Radial-Distance-U-Re-ID} & 93.6 & 97.2 & 98.3 & 81.6\\
    SSG \cite{SSG} & 86.2& 94.6 & 96.5 & 68.7\\
    \hline
    U-SSL(GoogLeNet) & 86.9 & 95.1 & 96.6 & 62.2\\
    U-SSL(ResNet-50) & 88.6 & 95.2 & 96.8 & 68.7\\
    U-SSL(GoogLeNet*) & 89.7 & 96.4 & 98.8 & 74.7\\
    U-SSL(ResNet-50*) & \textbf{94.1} & \textbf{97.4} & \textbf{98.8} & \textbf{82.3}\\
  \hline
\end{tabular}
\end{table}

\begin{table}[t]
  \caption{Comparison with SOTA on the CUHK03-NP dataset. U-SSL (*) means that we densely concatenate the features from each feature block. The best results are in boldface.}
  \label{tab:CUHK03-reID-CMC}
  \center
  \begin{tabular}{|r|cc|cc|}
    \hline
   Model & \multicolumn{2}{c|}{Labeled} & \multicolumn{2}{c|}{Detected}\\
   \cline{2-5}
      & Rank-1 & mAP & Rank-1 & mAP\\
    \hline
    CAMEL \cite{CAMEL} & - & - & 31.9 & -\\
    PTGAN \cite{PTGAN} & 37.5 & - &-&-\\
    MLFN \cite{MLFN} & 54.7 & 49.2 & 52.8 & 47.8 \\
    DaRe(R)+RE+RR \cite{DaRe} &72.9 & 73.7 & 69.8 & 71.2\\
    DG-Net \cite{DG-Net} & -& -&65.6 & 61.1\\
    \hline
    U-SSL(GoogLeNet) & 77.1 & 75.7 & 70.7 & 69.1\\
    U-SSL(ResNet-50) & 78.9 & 78.2 & 72.5 & 69.6 \\
    U-SSL(GoogLeNet*) & 79.6 & 78.9 & 72.6 & 69.2\\
    U-SSL(ResNet-50*) & \textbf{79.9} & \textbf{80.4} & \textbf{74.8} & \textbf{69.7}\\
  \hline
\end{tabular}
\end{table}


\begin{table}[t]
  \caption{Comparison with SOTA on the MSMT17 dataset. U-SSL (*) means that we densely concatenate the features from each feature block. The best results are in boldface.}\label{tab:MSMT17-CMC}
  \center
  \begin{tabular}{|r|ccc|c|}
    \hline
    Model  & Rank-1 & Rank-5 & Rank-10 & mAP\\
    \hline
    UST \cite{Unsupervised-TIP20}  & 31.4 & 41.4 & 45.7 & 9.9\\
    Exemplar \cite{Exemplar} & 30.2 & 41.5 & 46.8 & 10.2\\
    \hline
    GLAD \cite{GLAD} & 61.4 &76.8 &81.6& 34.0 \\
    PDC \cite{PDC} & 58.0 & 73.6 & 79.4& 29.7 \\
    MMCL \cite{MMCL} & 35.4 & 44.8 & 49.8 & 11.2\\
    MMT \cite{MMT} & 58.8 &71.0 &76.1 & 29.7\\
    Self-paced-CL \cite{Self-paced-CL} & 42.3 & 55.6 & 61.2 & 19.1\\
    PT-RDC \cite{Radial-Distance-U-Re-ID} & 69.9 & 80.3 & 85.4 & 40.7\\
    \hline
    U-SSL(GoogLeNet) & 69.4 & \textbf{82.1} & \textbf{85.5} & 39.2 \\
    U-SSL(ResNet-50) & \textbf{71.1} & \textbf{83.3} & \textbf{87.0} & \textbf{40.9} \\
    U-SSL(GoogLeNet*) & \textbf{70.2} & \textbf{87.9} & \textbf{89.6} & \textbf{41.7}\\
    U-SSL(ResNet-50*) & \textbf{73.2} & \textbf{89.4} & \textbf{90.8} & \textbf{43.1}\\
  \hline
\end{tabular}
\end{table}

\subsection{Ablation Studies}\label{ssec:ablation}
In this section, we perform ablation studies to gain insights into the proposed U-SSL. We first investigate the choice of surrogate classes with different class numbers and the number of samples per surrogate class. Then, we study the effect of gradient-guided similarity separation as well as the role of self-similarity learning in our method.

\paragraph{Effect of Surrogate Classes}

\begin{figure}[t]
  \begin{tabular}{cc}
      \hspace{-0.5cm}  \includegraphics[width=0.5\linewidth]{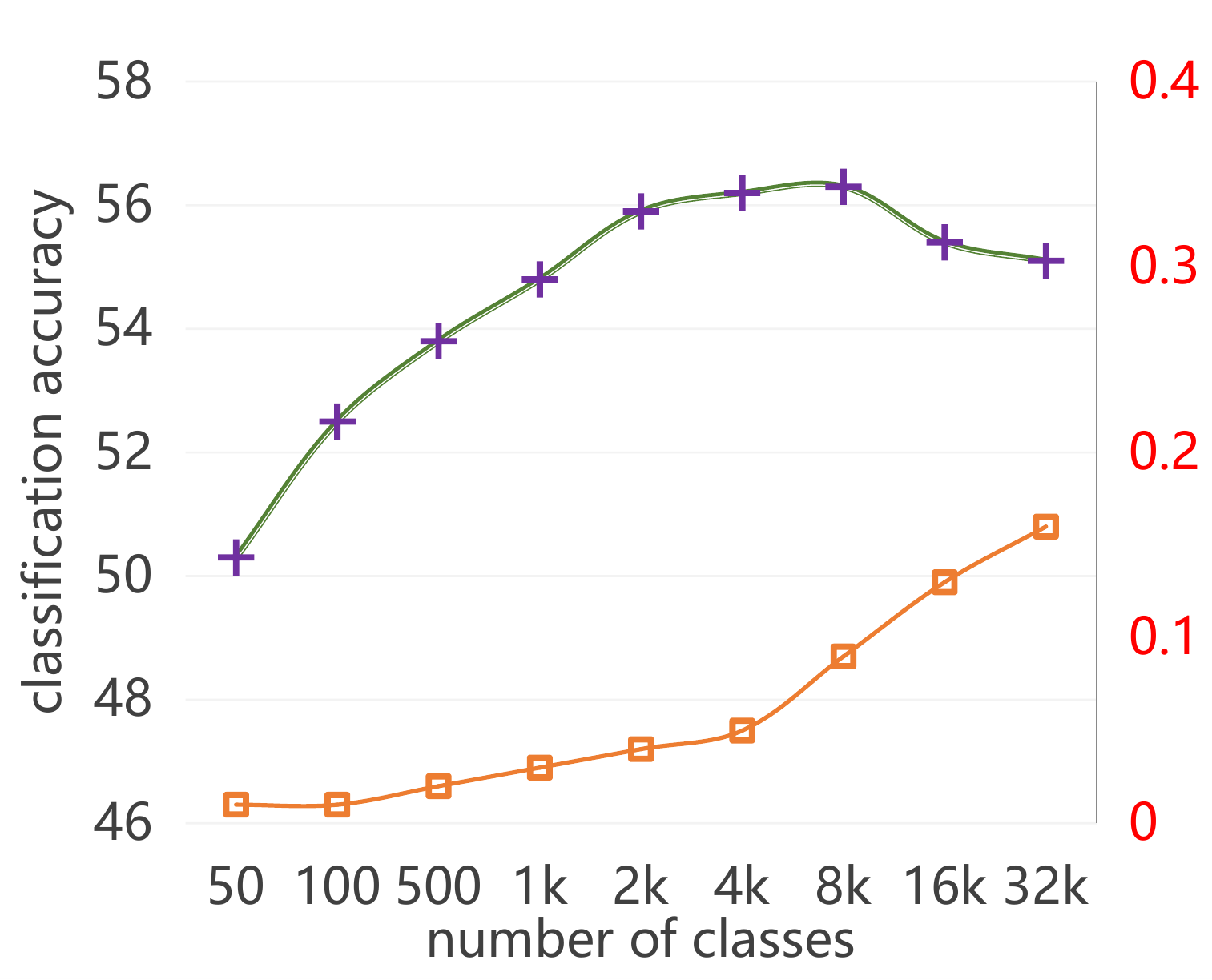}
       &\includegraphics[width=0.5\linewidth]{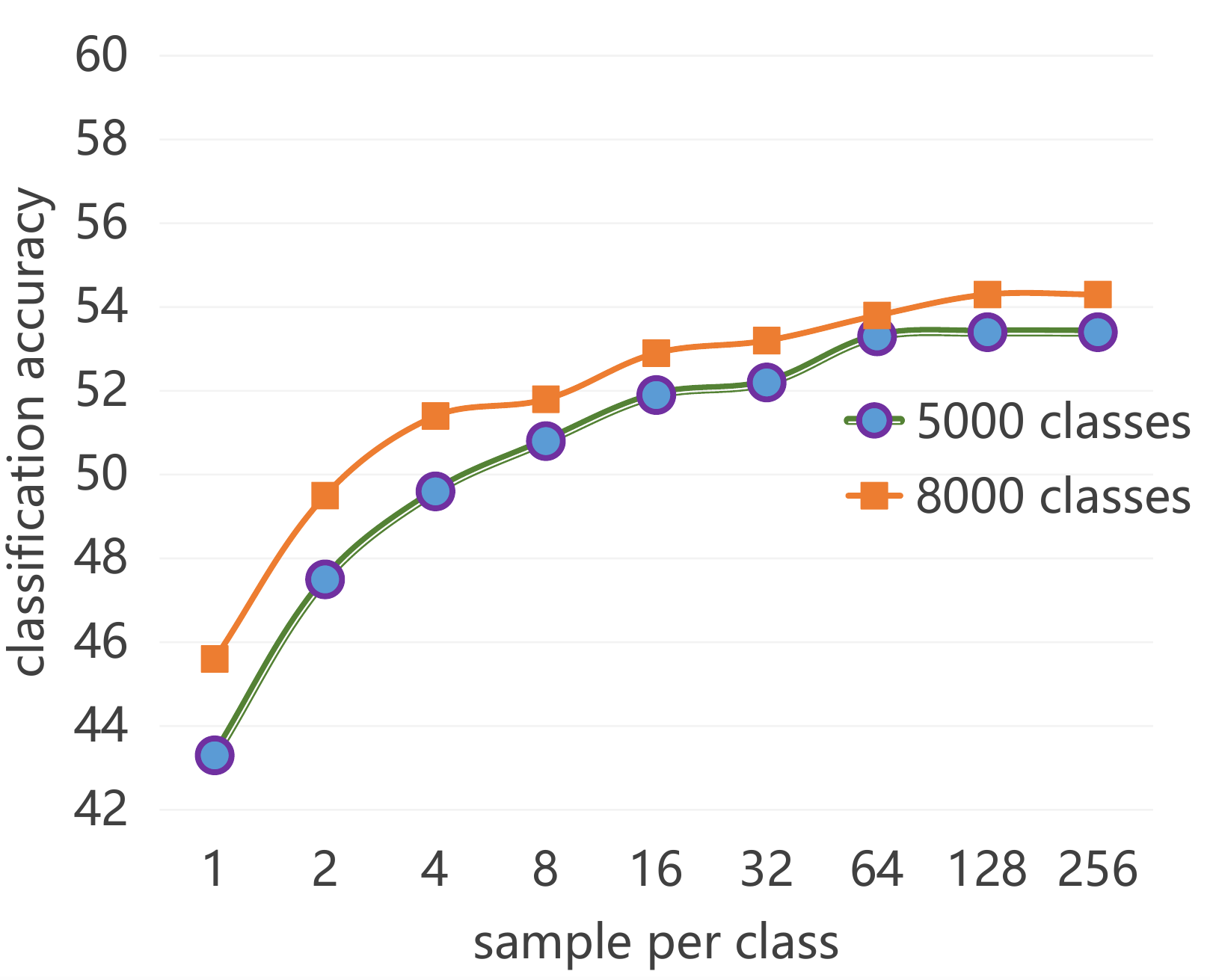}
  \end{tabular}
  \caption{Influence of the number of surrogate classes. The validation error on the surrogate data is plotted in red. Note we have different $y$-axes for the two curves.}\label{fig:number-class}
\end{figure}

To directly evaluate the choice of surrogate classes, we conduct a classification task to discriminate these classes, where we adopt the ResNet-50 \cite{Resnet} as the backbone and extract the components of the network before the soft-max layer. The number of surrogate classes, i.e., $N$ varies between 50 and 32,000. We conduct the trial on Market-1501 dataset, which is suitable to patch sampling because the dataset contains a moderately even-numbered of images regarding each identity. The results are shown in Fig. \ref{fig:number-class} (a). It clearly shows that the classification accuracy increases as more surrogate classes are included in the training. It reaches an optimum at around 8,000 classes and starts decreasing thereafter. This is because the larger the number of surrogate classes, the more likely it is to draw similar samples close to a group. However, when the number of surrogate classes increases, collisions may undermine the set of surrogate labels because the classification tends to wrongly classify the same patch into multiple labels. To prove that, we examine the validation error on the same dataset. The validation set is created by randomly choosing 100 disjoint identities from the training identities. It implies that oversampling on similar images may unexpectedly decrease the performance of the model. However, this drawback can be mitigated by our proposed gradient-guided similarity separation, which relates the same identity images into the same class.

Fig. \ref{fig:number-class} (b) shows the classification accuracy on the Market-1501 dataset when the number of training samples per surrogate class, i.e., $K$ varies between 1 and 256. The classification performance steadily increases with more samples per surrogate class and saturates at around 64 samples. This empirically indicates that creating 64 samples for each surrogate class is sufficient to approximate the class distinction. 

\paragraph{Effect of Gradient-Guided Similarity Separation}

\begin{figure}[t]
  \begin{tabular}{cc}
        \includegraphics[width=0.5\linewidth]{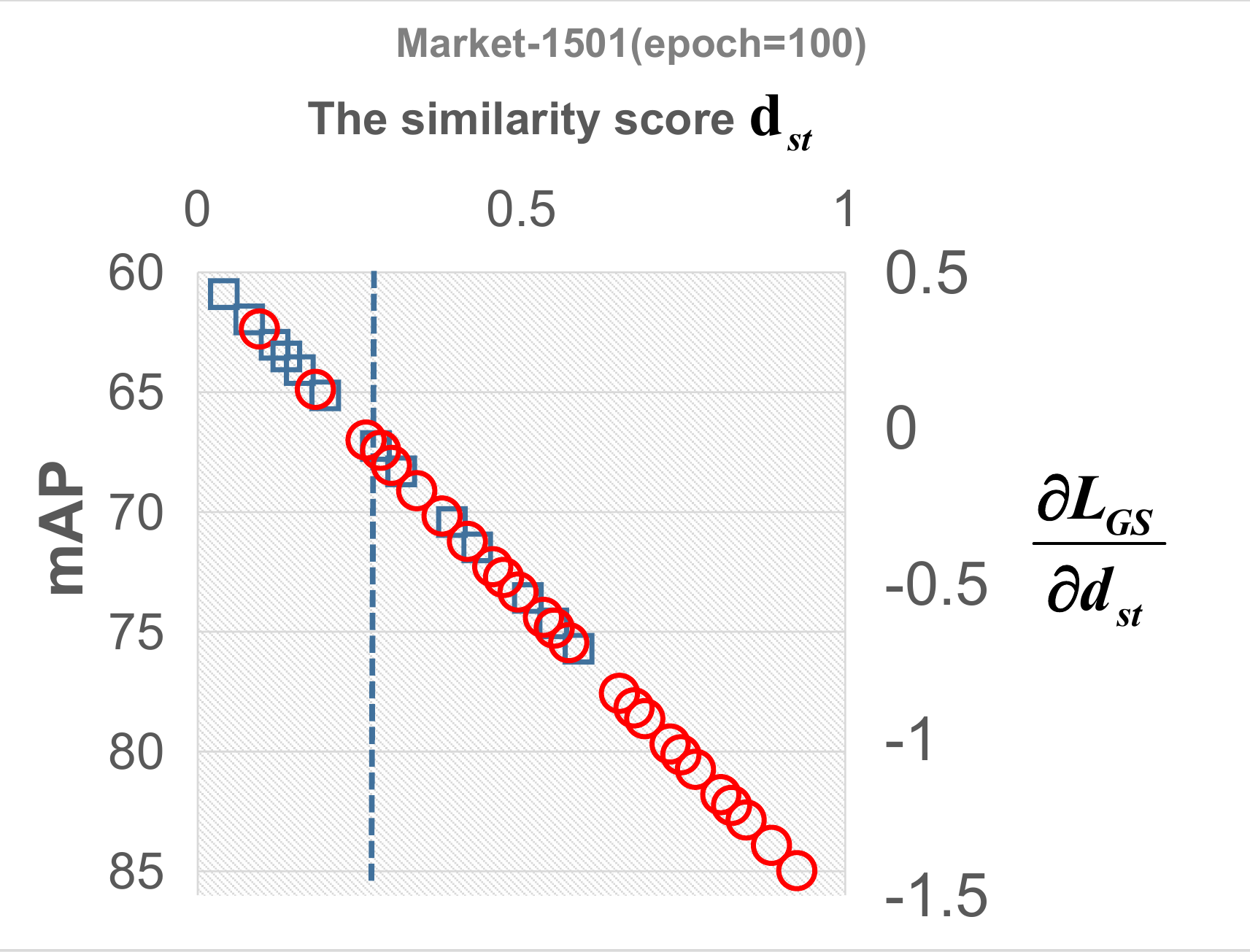}
       & \includegraphics[width=0.5\linewidth]{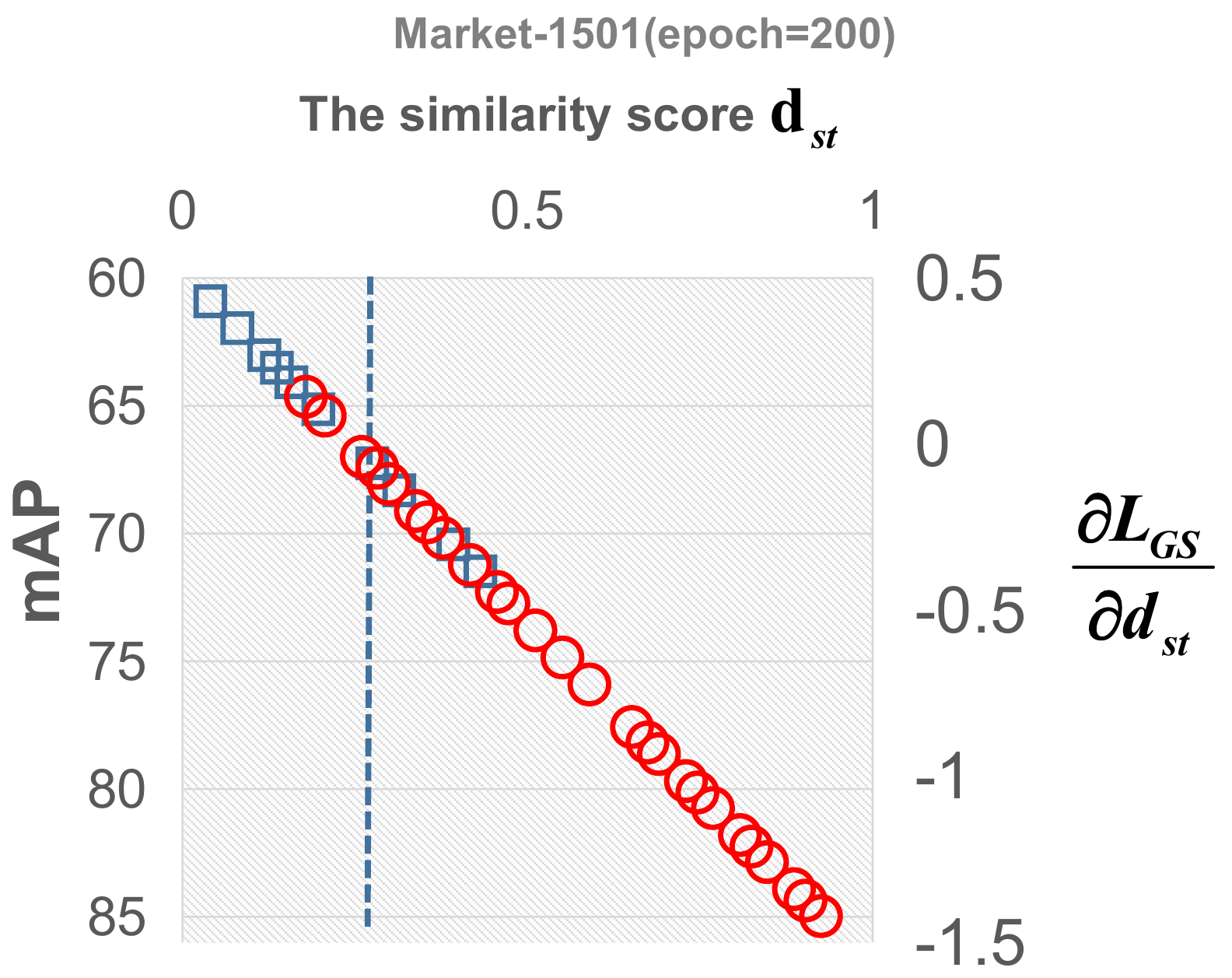}
  \end{tabular}
  \caption{Impact of training with the $\mathcal{L}_{GS}$ loss on the Market-1501 dataset. In each subfigure, the two axes show the mAP values at different stages and the gradient $\frac{\partial L_{GS}}{\partial d_{st}}|_{\theta} = 1- \alpha (d_{st}-\delta) $against the similarity scores $d_{st}$ between the query $\mathbf{x}_s$ and the gallery $\mathbf{x}_t$. Here, we empirically set $\alpha=2$ and $\delta=0.25$ is plotted as a vertical dashed line. Scores that are relevant (irrelevant) to the query images are visualized in the red circles (blue squares).}\label{fig:gradient-similarity}
\end{figure}

In this experiment, we evaluate the influence of the pairwise alignment between gradient vectors, and show the results in Fig. \ref{fig:gradient-similarity}. To do so, we select a query from the Market-1501 dataset and introspect the stages of epoch 100 to epoch 200. The gradients are plotted against the similarity scores $d_{st}$ between the query $\mathbf{x}_d$ and the gallery image $\mathbf{x}_t$, and scores for relevant/irrelevant images are colored in red and blue, respectively. For each query, the mAP values are also computed at each stage. As can be seen from the figure, the similarity scores for the query set are reasonably well separated as the training progresses into more epochs, so as to expedite the separation on relevant/irrelevant examples into increasingly tighter clusters. At the same time, most relevant examples are shown to be above the $\delta$ threshold. We conclude that when the similarity scores above $\delta$ are increased, the corresponding vectors will be pushed closer.

To show the relative role of $\mathcal{L}_{GS}$ in the overall objective optimization $\mathbb L$, we train four variants of our model by optimizing $\epsilon \cdot \mathcal{L}_{GS}$, where the coefficient $\epsilon$ empirically manipulates the role of $\mathcal{L}_{GS}$. Results are shown in Table \ref{tab:GS-effect}. When a similarity function optimizes $\mathcal{L}_{GS}$ alone, it cannot reduce the view difference, and therefore leads to a less effective metric. When the objective function Eq. \eqref{eq:obj-function} is combined with $\mathcal{L}_{GS}$ with the balance parameter empirically adjusted in the range of $\epsilon = \{0.001,0.01,1\}$, the model tends to learn the separation and feature mapping from semantic embeddings effectively. Finally, when the role of $\mathcal{L}_{GS}$ is fully incorporated into the objective function, the performance in terms of mAP and rank-1 peaks on two datasets.

\begin{figure}[t]
  \begin{tabular}{cc}
  \hspace{-0.5cm}    \includegraphics[width=0.5\linewidth]{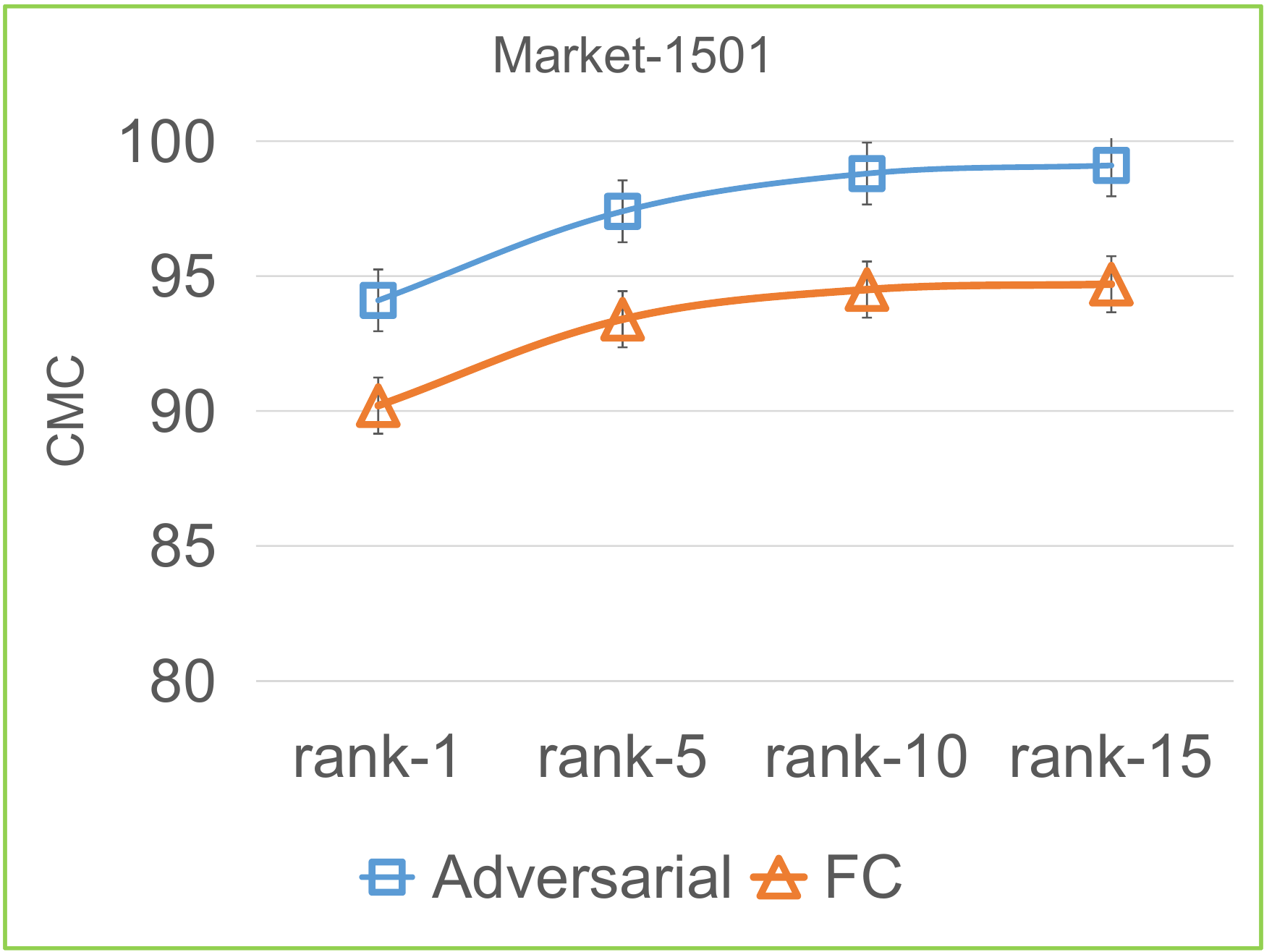}
       & \includegraphics[width=0.5\linewidth]{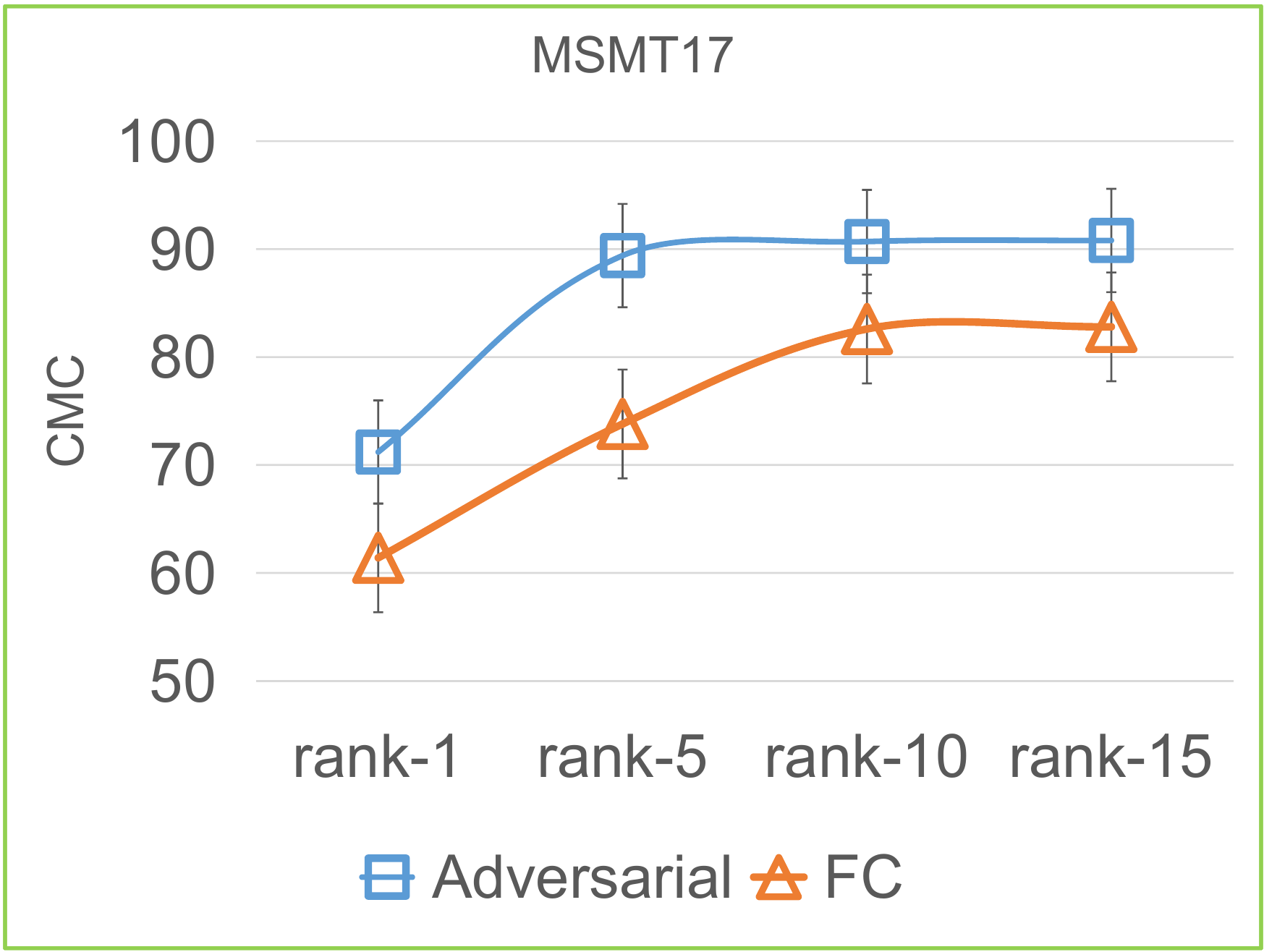}
  \end{tabular}
  \caption{Comparison results of adversarial training and the use of fully connection in intra-similarity learning.}\label{fig:Adv-vs-FC}
\end{figure}

\begin{figure}[t]
  \begin{tabular}{cc}
     \hspace{-0.5cm}   \includegraphics[width=0.5\linewidth]{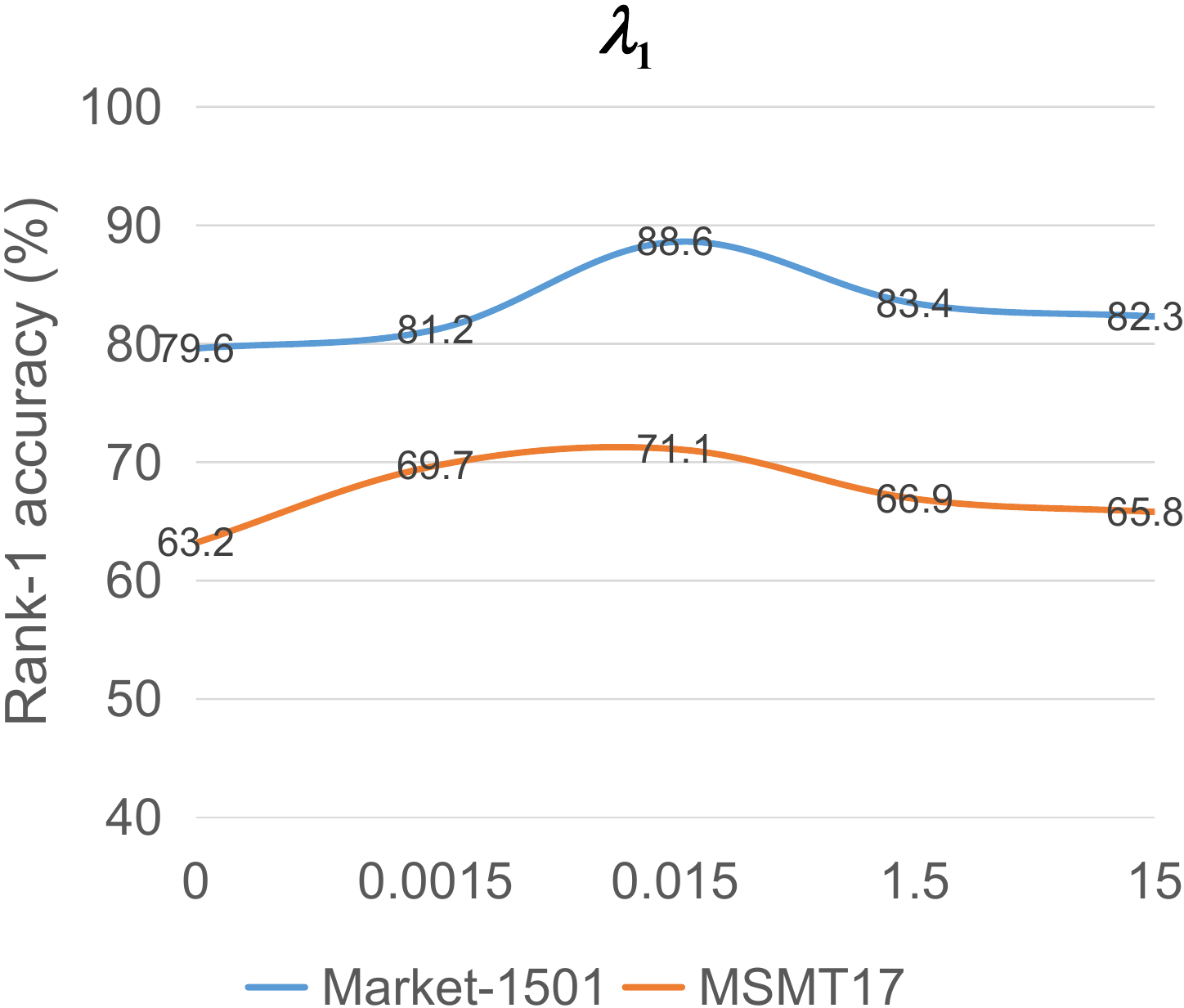}
       & \includegraphics[width=0.5\linewidth]{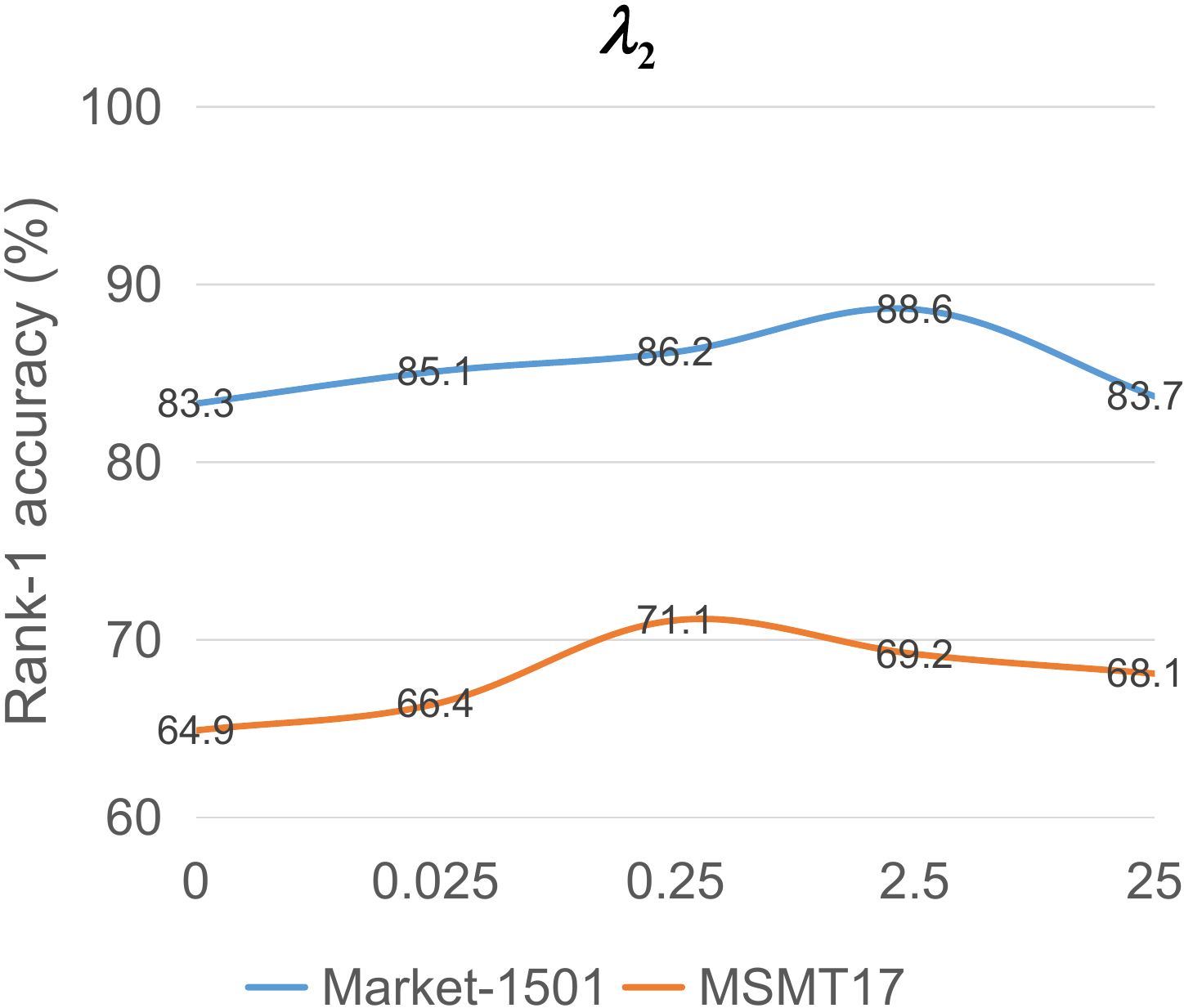}
  \end{tabular}
  \caption{Evaluation of hyper-parameters $\lambda_1$ and $\lambda_2$.}\label{fig:lambda}
\end{figure}

\begin{table}[t]
\caption{Effect of gradient-guided separation on two datasets.}
\begin{center}
\begin{tabular}{|r|cc|cc|}
    \hline
    \multirow{2}{*}{Model} & \multicolumn{2}{c|}{Market-1501} & \multicolumn{2}{c|}{MSMT17}\\
    \cline{2-5}
     & mAP & Rank-1 & mAP & Rank-1\\
     \hline
     $(\epsilon = 0.001) \mathcal{L}_{GS} \rightarrow \mathbb L$ & 34.1 & 89.2 & 38.9 & 63.5\\
     $ (\epsilon=0.01) \mathcal{L}_{GS} \rightarrow \mathbb L$ & 38.4 & 91.4 & 41.7 & 68.6\\
     $ (\epsilon=1) \mathcal{L}_{GS} \rightarrow \mathbb L$ & \textbf{82.3} & \textbf{94.1} & \textbf{43.1} & \textbf{71.2}\\
    \hline
\end{tabular}
\end{center}
\label{tab:GS-effect}
\end{table}

\begin{table}[t]
\caption{Effect of intra-similarity learning on the optimization.}
\begin{center}
\begin{tabular}{|r|cc|cc|}
    \hline
    Model & \multicolumn{2}{c|}{Market-1501} & \multicolumn{2}{c|}{MSMT17}\\
    \cline{2-5}
 $\lambda_1 \mathcal{L}_{Adv}$+$\lambda_2 \mathcal{L}_S$& mAP & Rank-1 & mAP & Rank-1\\
     \hline
     $(\lambda_1=0,\lambda_2=0.25) $ & 75.7 & 86.6& 34.8 & 55.4\\
     $(\lambda_1=15,\lambda_2=25) $ & 78.8 & 89.2 & 36.4 & 56.1\\
     $(\lambda_1=0.15,\lambda_2=2.5) $ & 79.2 & 90.7& 37.8 & 57.8\\
     $(\lambda_1=0.015,\lambda_2=0.25)$ & \textbf{82.3} & \textbf{94.1} & \textbf{43.1} & \textbf{71.2}\\
    \hline
\end{tabular}
\end{center}
\label{tab:div-effect}
\end{table}

\begin{table}[t]
  \caption{Rank-1 value (mAP) for our model with/without the non-local blocks.}
  \label{tab:with-NB}
  \center
  \begin{tabular}{|lccc|}
    \hline
    Model &  Market-1501 & CUHK03-NP & MSMT17\\
    \hline
    w/o NB & 88.2 (77.4) & 0.0 (0.0) & 87.1 (38.7)\\
    w NB & \textbf{94.1 (82.3)} & \textbf{0.0 (0.0)} & \textbf{71.2 (43.1)} \\
  \hline
\end{tabular}
\end{table}

\paragraph{ Effect of Intra-Similarity Learning}

In the self-similarity learning paradigm, the intra-similarity learning is deployed to detect body parts of identities. Then, the resulting responses are fed to the cross-image inter-similarity learning to encode the similarity estimation. In this experiment, we replace the adversarial learning by employing a spatial attention to integrate the channels of the CPAMs. More specifically, for each CPAM$[j]=\hat{U}$ corresponding to the $j$-th attribute, we first perform an average pooling to aggregate spatial features into a channel signature: $\mathbb U =\frac{1}{h\times w}\sum_{i=1}^h \sum_{j=1}^w \hat{U}_{i,j,1:C}$. Then, we add two fully connected layers to the resultant signature as $ReLU(\mathbf W_2 \times ReLU(\mathbf W_1 \mathbb U))$. The comparison results on two datasets are given in Fig. \ref{fig:Adv-vs-FC}. The adversarial constraint on the channels clearly leads to higher accuracy values compared to the simplified average feature aggregation plus the fully connected layers. We further evaluate the contribution of intra-similarity in the whole optimization, and the results on two datasets are shown in Table \ref{tab:div-effect}. We remark that the essence of feature diversity is ensured by the min-max game of the CPAMs and SA learners. Thus, we fix the parameter $\lambda_2$ and adjust $\lambda_1$ to show the effect of $\mathcal{L}_{Adv}$ in the whole optimization. Table \ref{tab:div-effect} shows that $\lambda_1$ can be empirically set to a smaller value, i.e., $\lambda_1=0.015$, which entails that more enforcement should be imposed on $\mathcal{L}_{adv}$ because the resultant local features from the intra-similarity learning are vital to the inter-similarity learning.

\begin{table*}[t]
\caption{Comparison with SOTA methods on unsupervised domain adaptation for person re-ID.}
\begin{center}
\begin{tabular}{|c|c|r|cccc|}
    \hline
    Source & Target & Method & Rank-1 & Rank-5 & Rank-10 & mAP\\
    \hline
    \multirow{6}{*}{Market-1501} & \multirow{6}{*}{MSMT17} & HM+Self-paced-CL \cite{Self-paced-CL} &53.7& 65.0& 69.8& 26.8\\ 
    && SSG \cite{SSG} & 31.6 &- & 49.6 & 13.2\\
    && PTGAN \cite{PTGAN} &10.2&-&24.2&2.9\\
    && MMCL \cite{MMCL} & 40.8&51.8& 56.7&15.1\\
    && MMT-DBSCAN \cite{MMT} &50.1 & 63.5 & 69.3 & 24.0\\
    && U-SSL+Self-paced-CL \cite{Self-paced-CL} & \textbf{56.3} & \textbf{69.1} & \textbf{73.8} & \textbf{30.2}\\
    \hline
    \multirow{6}{*}{MSMT17} & \multirow{6}{*}{Market-1501} & HM+Self-paced-CL \cite{Self-paced-CL} &89.7& 96.1& 97.6& 77.5\\ 
    && MAR \cite{MAR} & 67.7 & 81.9 &-&40.0\\
    && PAUL \cite{PAUL} & 68.5 & 82.4 & 87.4 & 40.1\\
    && CASCL \cite{CASCL} & 65.4 & 80.6 & 86.2 & 35.5\\
    && MMT-DBSCAN \cite{MMT} & 89.3& 95.8 & 97.5& 75.6\\
    && U-SSL+Self-paced-CL \cite{Self-paced-CL} & \textbf{91.2}& \textbf{97.1} & \textbf{98.0} & \textbf{79.1}\\
    \hline
\end{tabular}
\end{center}
\label{tab:generalization}
\end{table*}

\paragraph{Effect of The Non-local Block}

To validate the importance of the non-local block in the deformable convolution, we compare the performance of the proposed algorithm with and without non-local blocks. As shown in Table \ref{tab:with-NB}, the network with non-local blocks consistently achieves superior performance. This indicates that non-local blocks are helpful in capturing the long-range dependencies and the correlations of each feature, which are necessary to estimate the offset. Without the non-local blocks, we observe a performance drop of 4.7\% at the rank-1 value on Market-1501, and a drop by 3.8\%, 3.3\% on the other two datasets.

\paragraph{The Study on Hyper-Parameters}

In this experiment, we empirically evaluate the hyper-parameters $\lambda_1$ and $\lambda_2$ to determine their optimal values. Specifically, we vary the value of one parameter, e.g., $\lambda_1$, ranging from 0 to 100, while fixing the other. The evaluation results are shown in Fig. \ref{fig:lambda}. We can see that the empirically optimal values for the hyper-parameters are $\lambda_1=0.015$, $\lambda_2=2.5$. Thus, we use these values as default if not specified otherwise.

\paragraph{Generalization Ability}
We have evaluated how our proposed method generalizes from a source (training) dataset to a different target dataset. We adopted the training model on one dataset, e.g., Market-1501, and then adapted the model to a different target dataset, e.g., MSMT17. We split the target-domain data into clusters and un-clustered outliers by using the DBSCAN algorithm \cite{DBSCAN}. To adapt the source domain to the target domain in an unsupervised manner, we adopt the self-paced learning \cite{Self-paced-CL}, where in the re-clustering step before each epoch, only the most reliable clusters are preserved while the unreliable clusters are disassembled back to un-clustered instances. Following \cite{Self-paced-CL}, the unreliable clusters can be identified by measuring the independence and compactness. Obtained experimental results are reported in Table \ref{tab:generalization}, which show that our method performs better than the SOTA methods. This is mainly due to the reliable clusters constructed by our gradient-guided similarity algorithm. Our results demonstrate that the clustering reliability impacts the learned representation. This observation is consistent with HM-Self-paced-CL \cite{Self-paced-CL}.

%% file: Conclusion.tex
\section{Conclusions}\label{sec:con}

In this paper, we propose a novel unsupervised self-similarity learning paradigm for person re-identification. We reformulate the pseudo labelling as the trainable surrogate classes and generate images in pseudo pairs by measuring pairwise gradient alignments. This scheme can effectively improve the pseudo labeling with relative pairwise comparison, instead of accessing the global clustering structure. To enhance the generalization of the learned metric towards unseen shots, self-similarity learning is performed to identify diverse patches. This is achieved through adversarial training to obtain local discriminatives feature from individual images, namely intra-similarity learning. Meanwhile, we promote the inter-similarity learning across images via a deformable convolution with non-local block connections. We demonstrate the superiority of our method on several benchmark datasets.

%% file: bare_jrnl_compsoc.bbl
\begin{thebibliography}{10}
\providecommand{\url}[1]{#1}
\csname url@samestyle\endcsname
\providecommand{\newblock}{\relax}
\providecommand{\bibinfo}[2]{#2}
\providecommand{\BIBentrySTDinterwordspacing}{\spaceskip=0pt\relax}
\providecommand{\BIBentryALTinterwordstretchfactor}{4}
\providecommand{\BIBentryALTinterwordspacing}{\spaceskip=\fontdimen2\font plus
\BIBentryALTinterwordstretchfactor\fontdimen3\font minus
  \fontdimen4\font\relax}
\providecommand{\BIBforeignlanguage}[2]{{%
\expandafter\ifx\csname l@#1\endcsname\relax
\typeout{** WARNING: IEEEtran.bst: No hyphenation pattern has been}%
\typeout{** loaded for the language `#1'. Using the pattern for}%
\typeout{** the default language instead.}%
\else
\language=\csname l@#1\endcsname
\fi
#2}}
\providecommand{\BIBdecl}{\relax}
\BIBdecl

\bibitem{CAN}
H.~Liu, J.~Feng, M.~Qi, J.~Jiang, and S.~Yan, ``End-to-end comparative
  attention networks for person re-identification,'' \emph{IEEE Transactions on
  Image Processing}, vol.~26, no.~7, pp. 3492--3506, 2017.

\bibitem{Part-Bilinear}
Y.~Suh, J.~Wang, S.~Tang, T.~Mei, and K.~M. Lee, ``Part-aligned bilinear
  representations for person re-identification,'' in \emph{ECCV}, 2018.

\bibitem{Improved-re-ID}
E.~Ahmed, M.~Jones, and T.~K. Marks, ``An improved deep learning architecture
  for person re-identification,'' in \emph{CVPR}, 2015.

\bibitem{Dual-part-align}
J.~Guo, Y.~Yuan, L.~Huang, C.~Zhang, J.-G. Yao, and K.~Han, ``Beyond human
  parts: Dual part-aligned representations for person re-identification,'' in
  \emph{ICCV}, 2019.

\bibitem{CASCL}
A.~Wu, W.-S. Zheng, and J.-H. Lai, ``Unsupervised person re-identification by
  camera-aware similarity consistency learning,'' in \emph{ICCV}, 2019, pp.
  6922--6931.

\bibitem{Deep-re-ID}
W.~Li, R.~Zhao, T.~Xiao, and X.~Wang, ``Deep-reid: Deep filter pairing neural
  network for person re-identification,'' in \emph{CVPR}, 2014.

\bibitem{Defence-triplet-loss}
A.~Hermans, L.~Beyer, and B.~Leibe, ``In defence of the triplet loss for person
  re-identification,'' in \emph{arXiv:1703.07737}, 2017.

\bibitem{Chen-quadruplet-cvpr2017}
W.~Chen, X.~Chen, J.~Zhang, and K.~Huang, ``Beyond triplet loss: a deep
  quadruplet network for person re-identification,'' in \emph{CVPR}, 2017.

\bibitem{Wu-CVIU-2018}
L.~Wu, Y.~Wang, Z.~Ge, Q.~Hu, and X.~Li, ``Structured deep hashing with
  convolutional neural networks for fast person re-identification,''
  \emph{Computer Vision and Image Understanding}, vol. 167, pp. 63--73, 2018.

\bibitem{Wu-TCSVT}
L.~Wu, R.~Hong, Y.~Wang, and M.~Wang, ``Cross-entropy adversarial view
  adaptation for person re-identification,'' \emph{IEEE Transactions on
  Circuits and Systems for Video Technology}, vol.~30, no.~7, pp. 2081--2020,
  2020.

\bibitem{WU-Video-Re-ID}
L.~Wu, Y.~Wang, H.~Yin, M.~Wang, and L.~Shao, ``Few-shot deep adversarial
  learning for video-based person re-identification,'' \emph{IEEE Transactions
  on Image Processing}, vol.~29, no.~1, pp. 1233--1245, 2020.

\bibitem{SPGAN}
W.~Deng, L.~Zheng, Q.~Ye, G.~Kang, Y.~Yang, and J.~Jiao, ``Image-image domain
  adaptation with preserved self-similarity and domain-dissimilarity for person
  re-identification,'' in \emph{CVPR}, 2018.

\bibitem{PTGAN}
L.~Wei, S.~Zhang, W.~Gao, and Q.~Tian, ``Person transfer gan to bridge domain
  gap for person re-identification,'' in \emph{CVPR}, 2018.

\bibitem{Exemplar}
Z.~Zhong, L.~Zheng, Z.~Luo, S.~Li, and Y.~Yang, ``Invariance matters: Exemplar
  memory for domain adaptive person re-identification,'' in \emph{CVPR}, 2019,
  pp. 598--607.

\bibitem{MMT}
Y.~Ge, D.~Chen, and H.~Li, ``Mutual mean-teaching: Pseudo label refinery for
  unsupervised domain adaptation on person re-identification,'' in \emph{ICLR},
  2020.

\bibitem{BUC}
Y.~Lin, X.~Dong, L.~Zheng, Y.~Yan, and Y.~Yang, ``A bottom-up clustering
  approach to unsupervised person re-identification,'' in \emph{AAAI}, 2019.

\bibitem{Unsupervised-TIP20}
Y.~Lin, Y.~Wu, C.~Yan, M.~Xu, and Y.~Yang, ``Unsupervised person
  re-identification via cross-camera similarity exploration,'' \emph{IEEE
  Transactions on Image Processing}, vol.~29, pp. 5481--5490, April 2020.

\bibitem{MMCL}
D.~Wang and S.~Zhang, ``Unsupervised person re-identification via multi-label
  classification,'' in \emph{CVPR}, 2020, pp. 10\,981--10\,990.

\bibitem{SSL}
Y.~Lin, L.~Xie, Y.~Wu, C.~Yan, and Q.~Tian, ``Unsupervised person
  re-identification via softened similarity learning,'' in \emph{CVPR}, 2020,
  pp. 3390--3399.

\bibitem{TJ-AIDL}
J.~Wang, X.~Zhu, S.~Gong, and W.~Li, ``Transferable joint attribute-identity
  deep learning for unsupervised person re-identification,'' in \emph{CVPR},
  2018, pp. 2275--2284.

\bibitem{PAUL}
Q.~Yang, H.-X. Yu, A.~Wu, and W.-S. Zheng, ``Patch-based discriminative feature
  learning for unsupervised person re-identification,'' in \emph{CVPR}, 2019,
  pp. 3633--3642.

\bibitem{Radial-Distance-U-Re-ID}
S.~Seth, A.~Sonth, and A.~Chakraborty, ``Pose-transformation and radial
  distance clustering for unsupervised person re-identification,'' in
  \emph{BMVC}, 2021.

\bibitem{Exemplar-CNN}
A.~Dosovitskiy, P.~Fischer, J.~T. Springenberg, M.~Riedmiller, and T.~Brox,
  ``Discriminative unsupervised feature learning with exemplar convolutional
  neural networks,'' \emph{IEEE Transactions on Pattern Analysis and Machine
  Intelligence}, vol.~38, no.~9, pp. 1734--1747, 2016.

\bibitem{SSG}
Y.~Fu, Y.~Wei, G.~Wang, Y.~Zhou, H.~Shi, and T.~S. Huang, ``Self-similarity
  grouping: A simple unsupervised cross domain adaptation approach for person
  re-identification,'' in \emph{ICCV}, 2019, pp. 6113--6121.

\bibitem{Deform-Conv}
J.~Dai, H.~Qi, Y.~Xiong, Y.~Li, G.~Zhang, H.~Hu, and Y.~Wei, ``Deformable
  convolutional networks,'' in \emph{ICCV}, 2017.

\bibitem{Non-local}
X.~Wang, R.~Girshick, A.~Gupta, and K.~He, ``Non-local neural networks,'' in
  \emph{CVPR}, 2018.

\bibitem{SSEN}
G.~Shim, J.~Park, and I.~S. Kweon, ``Robust reference-based super-resolution
  with similarity-aware deformable convolution,'' in \emph{CVPR}, 2020, pp.
  8425--8434.

\bibitem{LOMO}
S.~Liao, Y.~Hu, X.~Zhu, and S.~Z. Li, ``Person re-identification by local
  maximal occurrence representation and metric learning,'' in \emph{CVPR},
  2015.

\bibitem{Gaussian}
T.~Matsukawa, T.~Okabe, E.~Suzuki, and Y.~Sato, ``Hierarchical gaussian
  descriptor for person re-identification,'' in \emph{CVPR}, 2016.

\bibitem{U-Re-ID-salience-match}
R.~Zhao, W.~Ouyang, and X.~Wang, ``Person re-identification by salience
  matching,'' in \emph{ICCV}, 2013.

\bibitem{U-Re-ID-salience}
------, ``Unsupervised salience learning for person re-identification,'' in
  \emph{CVPR}, 2013.

\bibitem{L1-graph}
E.~Kodirov, T.~Xiang, Z.~Fu, and S.~Gong, ``Person re-identification by
  unsupervised l1 graph learning,'' in \emph{ECCV}, 2016.

\bibitem{SyRI}
S.~Bak, P.~Carr, and J.-F. Lalonde, ``Domain adaptation through synthesis for
  unsupervised person re-identification,'' in \emph{ECCV}, 2018, pp.~--.

\bibitem{Self-paced-CL}
Y.~Ge, F.~Zhu, D.~Chen, R.~Zhao, and H.~Li, ``Self-paced contrastive learning
  with hybrid memory for domain adaptive object re-id,'' in \emph{NeurIPS},
  2020.

\bibitem{SDA-MMT}
Y.~Ge, S.~Yu, and D.~Chen, ``Improved mutual mean-teaching for unsupervised
  domain adaptive re-id,'' in \emph{ECCVW}, 2020.

\bibitem{VCFL}
F.~Liu and L.~Zhang, ``View confusion feature learning for person,'' in
  \emph{ICCV}, 2019, pp. 6639--6648.

\bibitem{Camera-style-re-ID}
Z.~Zhong, L.~Zheng, Z.~Zheng, S.~Li, and Y.~Yang, ``Camera style adaptation for
  person re-identification,'' in \emph{CVPR}, 2018.

\bibitem{Local-SS}
E.~Shechtman and M.~Irani, ``Matching local self-similarities across images and
  videos,'' in \emph{CVPR}, 2007.

\bibitem{SELFYNet-TSM}
H.~Kwon, M.~Kim, S.~Kwak, and M.~Cho, ``Learning self-similarity in space and
  time as generalized motion for video action recognition,'' in \emph{ICCV},
  2021, pp. 13\,065--13\,075.

\bibitem{SSS}
Z.~Chen, F.~Li, Y.~Quan, Y.~Xu, and H.~Ji, ``Deep texture recognition via
  exploiting cross-layer statistical self-similarity,'' in \emph{CVPR}, 2021,
  pp. 5231--5240.

\bibitem{MSCAN}
D.~Li, X.~Chen, Z.~Zhang, and K.~Huang, ``Learning deep context-aware features
  over body and latent parts for person re-identification,'' in \emph{CVPR},
  2017, pp. 383--393.

\bibitem{WU-Co-attention}
L.~Wu, Y.~Wang, J.~Gao, M.~Wang, Z.-J. Zha, and D.~Tao, ``Deep co-attention
  based comparators for relative representation learning in person
  re-identification,'' \emph{IEEE Transactions on Neural Networks and Learning
  Systems}, vol.~32, no.~2, pp. 722--735, 2021.

\bibitem{HA-CNN}
W.~Li, X.~Zhu, and S.~Gong, ``Harmonious attention network for person
  re-identification,'' in \emph{CVPR}, 2018, pp. 2285--2294.

\bibitem{Curriculum-sampling}
C.~Wang, Q.~Zhang, C.~Huang, W.~Liu, and X.~Wan, ``Mancs: A multi-task
  attentional network with curriculum sampling for person re-identification,''
  in \emph{ECCV}, 2018.

\bibitem{Compact-Bilinear}
Y.~Gao, O.~Beijbom, N.~Zhang, and T.~Darrell, ``Compact bilinear pooling,'' in
  \emph{CVPR}, 2016.

\bibitem{What-and-where}
L.~Wu, Y.~Wang, X.~Li, and J.~Gao, ``What-and-where to match: Deep spatially
  multiplicative integration networks for person re-identification,''
  \emph{Pattern Recognition}, vol.~76, pp. 727--738, 2018.

\bibitem{BAT-net}
P.~Fang, J.~Zhou, S.~K. Roy, L.~Petersson1, and M.~Harandi, ``Bilinear
  attention networks for person retrieval,'' in \emph{ICCV}, 2019, pp.
  8030--8039.

\bibitem{Unsupervised-deep-cluster}
J.~Xie, R.~Girshick, and A.~Farhadi, ``Unsupervised deep embedding for
  clustering analysis,'' in \emph{ICML}, 2016, pp.~--.

\bibitem{GLAD}
L.~Wei, S.~Zhang, H.~Yao, W.~Gao, and Q.~Tian, ``Glad: Global-local-alignment
  descriptor for pedestrian retrieval,'' in \emph{ACM International Conference
  on Multimedia}, 2017.

\bibitem{E-Metric}
H.~Shi, Y.~Yang, X.~Zhu, S.~Liao, Z.~Lei, W.~Zheng, and S.~Z. Li, ``Embedding
  deep metric for person re-identification: A study against large variations,''
  in \emph{ECCV}, 2016, pp. 732--748.

\bibitem{SSM}
S.~Bai, X.~Bai, and Q.~Tian, ``Scalable person re-identification on supervised
  smoothed manifold,'' in \emph{CVPR}, 2017, pp. 1281--1287.

\bibitem{LDA-Re-ID}
L.~Wu, C.~Shen, and A.~van~den Hengel, ``Deep linear discriminant analysis on
  fisher networks: A hybrid architecture for person re-identification,''
  \emph{Pattern Recognition}, vol.~65, pp. 238--250, 2017.

\bibitem{Spindle-net}
H.~Zhao, M.~Tian, S.~Sun, J.~Shao, J.~Yan, S.~Yi, X.~Wang, and X.~Tang,
  ``Spindlenet: Person re-identification with human body region guided feature
  decomposition and fusion,'' in \emph{CVPR}, 2017.

\bibitem{DeML}
B.~Chen and W.~Deng, ``Hybrid-attention based decoupled metric learning for
  zero-shot image retrieval,'' in \emph{CVPR}, 2019.

\bibitem{Resnet}
K.~He, X.~Zhang, S.~Ren, and J.~Sun, ``Deep residual learning for image
  recognition,'' in \emph{CVPR}, 2016, pp. 770--778.

\bibitem{GRL}
Y.~Ganin, E.~Ustinova, H.~Ajakan, P.~Germain, H.~Larochelle, F.~Laviolette,
  M.~Marchand, and V.~Lempitsky, ``Domain-adversarial training of neural
  networks,'' \emph{Journal of Machine Learning Research}, vol.~17, no.~1, pp.
  2096--2030, 2006.

\bibitem{T-DAN}
Y.~Tian, Y.~Zhang, Y.~Fu, and C.~Xu, ``Tdan: Temporally deformable alignment
  network for video super-resolution,'' in \emph{arXiv:1812.02898}, 2018.

\bibitem{Market-1501}
L.~Zheng, L.~Shen, L.~Tian, S.~Wang, J.~dong Wang, and Q.~Tian, ``Scalable
  person re-identification:a benchmark,'' in \emph{ICCV}, 2015.

\bibitem{k-reciprocal}
Z.~Zhong, L.~Zheng, D.~Cao, and S.~Li, ``Re-ranking person re-identification
  with k-reciprocal encoding,'' in \emph{ICCV}, 2017.

\bibitem{DPM}
P.~F. Felzenszwalb, R.~B. Girshick, D.~McAllester, and D.~Ramanan, ``Object
  detection with discriminatively trained part-based models,'' \emph{IEEE
  Transactions on Pattern Analysis and Machine Intelligence}, vol.~32, no.~9,
  pp. 1627--1645, 2010.

\bibitem{Faster-r-cnn}
S.~Ren, K.~he, R.~Girshick, and J.~Sun, ``Faster r-cnn: towards real-time
  object detection with region proposal networks,'' in \emph{NIPS}, 2015, pp.
  91--99.

\bibitem{Pytorch}
A.~Paszke, S.~Gross, S.~Chintala, G.~Chanan, E.~Yang, Z.~DeVito, Z.~Lin,
  A.~Desmaison, L.~Antiga, and A.~Lerer, ``Automatic differentiation in
  pytorch,'' in \emph{NIPS Workshop}, 2017.

\bibitem{Adv-ML}
B.~Chen and W.~Deng, ``Energy confused adversarial metric learning for
  zero-shot image retrieval and clustering,'' in \emph{AAAI}, 2019.

\bibitem{Google-net}
C.~Szegedy, W.~Liu, Y.~Jia, P.~Sermanet, S.~Reed, D.~Anguelov, D.~Erhan,
  V.~Vanhoucke, and A.~Robinovich, ``Going deeper with convolutions,'' in
  \emph{CVPR}, 2014.

\bibitem{ACRN}
A.~Schumann and R.~Stiefelhagen, ``Person re-identification by deep learning
  attribute-complementary information,'' in \emph{CVPR Workshop}, 2017, pp.
  20--28.

\bibitem{Random-Erase}
Z.~Zhong, L.~Zheng, G.~Kang, S.~Li, and Y.~Yang, ``Random erasing data
  augmentation,'' in \emph{arXiv:1708.04896}, 2017.

\bibitem{CAMEL}
H.-X. Yu, A.~Wu, and W.-S. Zheng, ``Cross-view asymmetric metric learning for
  unsupervised person re-identification,'' in \emph{ICCV}, 2017.

\bibitem{MAR}
H.-X. Yu, W.-S. Zheng, A.~Wu, X.~Guo, S.~Gong, and J.-H. Lai, ``Unsupervised
  person re-identification by soft multilabel learning,'' in \emph{CVPR}, 2019,
  pp. 2148--2157.

\bibitem{UMDL}
P.~Peng, T.~Xiang, Y.~Wang, M.~Pontil, S.~Gong, T.~Huang, and Y.~Tian,
  ``Unsupervised cross-dataset transfer learning for person
  re-identification,'' in \emph{CVPR}, 2016, pp. 1306--1315.

\bibitem{HHL}
Z.~Zhong, L.~Zheng, S.~Li, and Y.~Yang, ``Generalizing a person retrieval model
  hetero-and homogeneously,'' in \emph{ECCV}, 2018.

\bibitem{MLFN}
X.~Chang, T.~M. Hospedales, and T.~Xiang, ``Multi-level factorisation net for
  person re-identification,'' in \emph{CVPR}, 2018.

\bibitem{DaRe}
Y.~Wang, L.~Wang, Y.~You, X.~Zou, and V.~Chen, ``Resource aware person
  re-identification across multiple resolutions,'' in \emph{CVPR}, 2018.

\bibitem{DG-Net}
Z.~Zheng, X.~Yang, Z.~Yu, L.~Zheng, Y.~Yang, and J.~Kautz, ``Joint
  discriminative and generative learning for person re-identification,'' in
  \emph{CVPR}, 2019.

\bibitem{PDC}
C.~Su, J.~Li, S.~Zhang, J.~Xing, W.~Gao, and Q.~Tian, ``Pose-driven deep
  convolutional model for person re-identification,'' in \emph{ICCV}, 2017, pp.
  3960--3969.

\bibitem{DBSCAN}
M.~Ester, H.~Kriegel, J.~Sander, and X.~Xu, ``A density-based algorithm for
  discovering clusters in large spatial databases with noise,'' in \emph{KDD},
  1996, pp. 226--231.

\end{thebibliography}
